 \documentclass[preprint,3p,times,twocolumn]{elsarticle}

\usepackage{amssymb}
\usepackage{amsthm}
\usepackage{amsmath}
\usepackage{hyperref}
\usepackage{booktabs}

\usepackage{natbib}
\usepackage{morefloats}
\usepackage{caption}
\usepackage{subcaption}
\usepackage{comment}
\newcommand{\Cllr}{\textit{C}$\mathrm{_{llr}}\,$}

\pdfoutput=1

\journal{}

\begin{document}


\date{April 12, 2023}
\begin{frontmatter}



\title{Embedding Aggregation for Forensic Facial Comparison}


\author[1,2]{Rafael Oliveira Ribeiro}



\ead{rafael.ror@pf.gov.br}



\affiliation[1]{organization={Department of Computer Science, University of Brasilia},
    addressline={Campus Universitário Darcy Ribeiro}, 
    city={Brasília},
    postcode={70910-900}, 
    country={Brazil}}
\affiliation[2]{organization={National Institute of Criminalistics},
    addressline={SPO Lote 7, Ed. INC}, 
    city={Brasília},
    postcode={70610-902}, 
    country={Brazil}}

\author[3]{João C. Neves}
\author[4]{Arnout C. C. Ruifrok}
\affiliation[3]{organization={NOVA-LINCS, University of Beira Interior},
    addressline={R. Marquês de Ávila e Bolama}, 
    city={Covilhã},
    postcode={6201-001}, 
    state={Castelo Branco},
    country={Portugal}}

\author[1]{Flávio de Barros Vidal}
\affiliation[4]{organization={Netherlands Forensic Institute},
    addressline={Laan van Ypenburg 6}, 
    city={The Hague},
    postcode={2497 GB}, 
    country={The Netherlands}}

\cortext[cor1]{Corresponding author}

\begin{abstract}
In forensic facial comparison, questioned-source images are usually captured in uncontrolled environments, with non-uniform lighting, and from non-cooperative subjects. The poor quality of such material usually compromises their value as evidence in legal matters. On the other hand, in forensic casework, multiple images of the person of interest are usually available. In this paper, we propose to aggregate deep neural network embeddings from various images of the same person to improve performance in facial verification. We observe significant performance improvements, especially for very low-quality images. Further improvements are obtained by aggregating embeddings of more images and by applying quality-weighted aggregation. We demonstrate the benefits of this approach in forensic evaluation settings with the development and validation of score-based likelihood ratio systems and report improvements in \Cllr of up to 95\% (from 0.249 to 0.012) for CCTV images and of up to 96\% (from 0.083 to 0.003) for social media images. 

\end{abstract}




\end{frontmatter}



\section{Introduction}

The increasing number of indoor and outdoor surveillance cameras and the widespread availability of smartphones has raised the number of crimes in which the perpetrator's facial image is recorded. This fact has fostered the interest in using these data to uncover the perpetrator's identity \citep{ashby_value_2017}. When a suspect is presented, the analysis of morphological facial features is currently recommended as the standard approach for the forensic comparison of faces~\citep{FISWG_2019}. This process is usually executed manually by comparing a set of defined facial morphological features in the questioned-source image with those in the suspects’ images (known-source images) \citep{zeinstra_2018}. The evaluation of the findings from the morphological analysis is often summarized in a qualitative scale of posterior probability (e.g., ``it is highly likely that the two images belong to the same identity'') or using qualitative scales based on the Likelihood Ratio (LR) (e.g., ``the similarities and differences observed are more likely when considering the images as belonging to the same identity rather than when considering they belong to distinct identities'') \citep{enfsi_di_bpm, enfsi_evaluative_reporting}.

\begin{figure}[h!]
    \centering
    \includegraphics[width=\columnwidth]{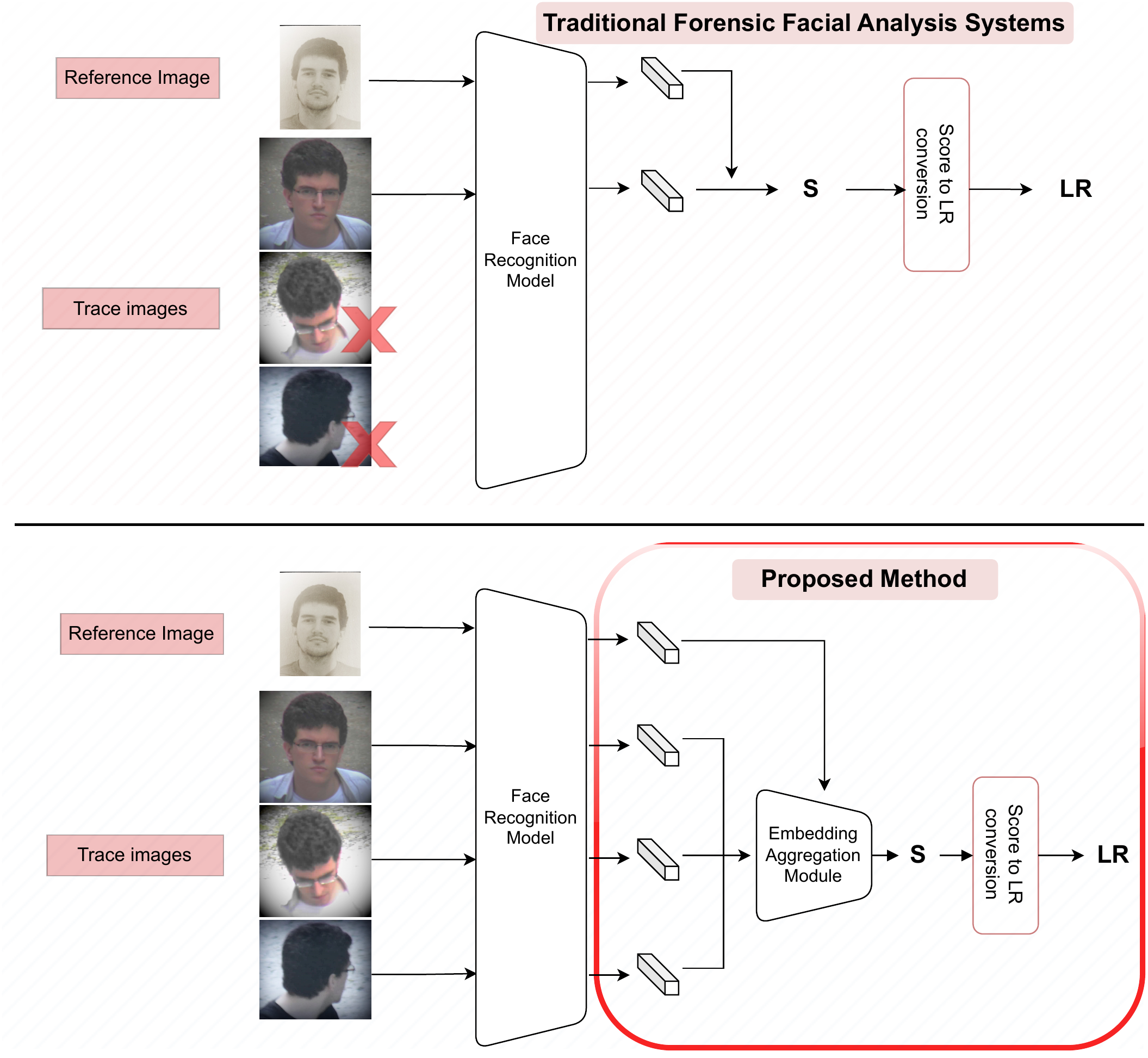}
    \caption{Comparison of the proposed framework with traditional forensic facial analysis systems.}
    \label{fig:intro}
\end{figure}

Although forensic practitioners using the current approach have demonstrated superior performance for facial comparisons relative to control groups \citep{Phillips2018_examiners_algorithms, hahn_2022_examiners_superrecognizers}, there has been a long-standing call for adopting more objective and quantitative methods in forensic science \citep{Saks2005, council_strengthening_2009, pcast_2016, Morrison2022}. In various fields related to biometric comparisons, the research community has responded to this call by investigating the possibility of using automated systems to quantify the evidence obtained from the data by computing an LR \citep{meuwly_2006_individualisation, neumann_statistical_2020, Brummer_CLLR, meuwly_2017_guideline_validation, morrison_consensus, RUIFROK2022111201_confusion_scores, ali_2014, jacquet, Molder_Leitet_2020_score_to_LR}. Based on the evaluation of comparison scores obtained from biometric samples, this new approach is especially appealing for the face modality for two reasons. Firstly, automatic facial recognition systems have experienced an enormous improvement in performance over the last few years \citep{frvt_2022, pereira_face_2022}. Secondly, the combined performance of human experts and facial recognition algorithms have been demonstrated to be superior to either the human experts or the algorithms alone \citep{phillips2014comparison, Phillips2018_examiners_algorithms}. 

Combining facial recognition systems’ outputs with human forensic examiners requires that both analyses are performed under the same evaluation paradigm. Currently, the LR paradigm is the recommended approach for evaluative reporting of source problems in forensic science \citep{enfsi_evaluative_reporting, nifs_aus_nz_eval_reporting}. Under this paradigm, forensic practitioners should express their evaluation using a likelihood ratio. The LR represents the degree of support of the evidence for one hypothesis relative to another mutually exclusive hypothesis. In this work, we consider common-source hypotheses, which, in the case of forensic facial comparison, are defined as:

\begin{enumerate}[\textbullet]
    \item $H_p$ (same-source hypothesis): Both the questioned-source and the known-source images depict the face of the same person; and;
    \item $H_d$ (different-source hypothesis): The questioned-source and the known-source image depict the faces of two different people from the same population\footnote{Often referred to as  \textit{reference population} in the forensic literature, it is the population from which an alternative suspect may have came from (e.g., young adult males from a specific region).}.
\end{enumerate}

The LR is computed according to

\begin{equation}
    LR = \frac{Pr(E|H_p, I)}{Pr(E|H_d, I)},
\label{eq:LR}
\end{equation}
i.e., it is the ratio of the probabilities ($Pr$) of obtaining the evidence ($E$) given each hypotheses and the contextual information ($I$) relevant to the case. From now on, we will omit $I$ in the equations, but the reader should remember that all probabilities related to the case are conditioned on $I$.

 Several works have proposed methods to obtain LRs by converting the scores of face recognition systems through the estimation of within-source and between-source distributions \citep{ali_2014, jacquet, Molder_Leitet_2020_score_to_LR}. However, the existing strategies focus only on a single questioned-source image, disregarding the possibility of aggregating information from multiple images (e.g., consecutive frames from CCTV footage) to compute a single LR. To address this limitation, as depicted in Figure~\ref{fig:intro}, we introduce a novel strategy for LR calculation in forensic facial comparison when multiple questioned-source images are available. The proposed method combines the facial descriptors\footnote{The facial descriptors in this work are the embeddings obtained from a Deep Convolution Neural Network-based facial recognition system.} of each sample to build into a single facial comparison score, which is subsequently mapped to an LR.
 
 The experiments performed in facial datasets representative of common forensic scenarios show that the proposed strategy decreases the log-likelihood ratio cost (\Cllr) compared to state-of-the-art face analysis approaches. Additionally, the proposed method is applicable even when the samples are obtained from non-consecutive moments in time, with varying illumination and pose.

The paper is organized as follows: in Section \ref{sec_related_work}, we review works on using biometric systems to evaluate faces as forensic evidence. In Section \ref{section_method}, the proposed method is described, and in Section \ref{section_data}, we detail the data used in this work. Section \ref{sec_experiments} describes the experiments performed, and Section \ref{sec_results} presents the results and discussion. We conclude in Section \ref{sec_conclusion}, presenting the limitations of this work and planned investigations on the same topic.

\section{Related Work}
\label{sec_related_work}
The possibility of using automated face recognition systems for quantifying forensic evidence has been studied for two decades~\citep{Ali_BTAS2013, Gonzalez-Rodriguez_FSI2005,Mandasari_IETB2014,jacquet,Molder_Leitet_2020_score_to_LR}. In 2005, Gonzalez-Rodriguez et al.~\citep{Gonzalez-Rodriguez_FSI2005} assessed the performance of face recognition approaches for forensic applications. The authors relied on a database comprising 295 identities. They used 400 within-source comparisons and 12,250 between-source comparisons for estimating the probability density functions of the two distributions, which were subsequently used to derive the LRs from similarity scores obtained from the recognition system. The improvement in face recognition accuracy and the development of more challenging datasets fostered the proposal of novel studies in the following years.

Ali \textit{et al.}~\citep{Ali_BTAS2013} evaluated the log-LR obtained from within-source and between-source scores of a commercial face recognition system. Nevertheless, the authors only analyzed five identities from the Face Recognition Grand Challenge (FRGC) dataset with 35 images per identity. Mandasari \textit{et al.}~\citep{Mandasari_IETB2014} introduced an innovative approach based on inter-session variability modeling followed by a linear transformation of the similarity score to obtain LRs of face recognition in the Surveillance Cameras Face Database (SCface), a database comprising samples from a usual forensic scenario. The first publicly available study using real forensic data was carried out by Mölder \textit{et al.}~\citep{Molder_WIFS2020}, which evaluated the effectiveness of the use of LRs obtained from facial comparison scores in forensic applications by using a national database of mugshots. However, few details were given concerning the face recognition algorithm, and the data used could not be shared.

In the last years, researchers have been proposing improvements to the traditional approach of inferring LRs from similarity scores obtained from recognition systems. Recently, Verma \textit{et al.}~\citep{Verma_FSIR2022} studied the performance of using LR for face verification using automatically detected facial landmarks for computing a set of morphometric facial indices, which were used as identity features. Despite having obtained an accuracy of 85\% when using $LR>1$ as the decision threshold, they only relied on a single dataset comprising 40 identities. Also, despite landmarks use being common in forensic scenarios~\citep{Porto_IJLM2020}, landmark-based approaches are highly dependent on the pose, which is unsuitable for CCTV footage. Anthropometric techniques are also not recommended for manually comparing face images for forensic evaluation \citep{fiswg_overview_2019}. Ruifrok \textit{et al.}~\citep{RUIFROK2022111201_confusion_scores} showed that the distribution of facial comparison scores could be used to assess the quality of trace images, which can be subsequently exploited to optimize the score-to-LR conversion, and consequently improve the discrimination and calibration of the obtained LRs. Zeinstra \textit{et al.}~\citep{Zeinstra_TIFS2017} analyzed the discrimination power of facial marks in forensic scenarios. The authors proposed an innovative method based on the number of marks in each cell of an auxiliary grid superimposed over the face. The number of marks along each cell is used as the facial features that are subsequently used by the face classifier. The evaluation of the \Cllr with respect to the number of facial marks and grid size evidenced the potential of this approach, even though the dataset considered is not particularly challenging for current face recognition systems regarding pose and occlusion.

\section{Proposed Method}
\label{section_method}

The comparison between a reference image ($X^r$) and a trace image ($X^t$) and the calibration of the resulting score $s$ into an $LR$ is the traditional strategy for quantifying evidence in forensic scenarios. The biometric score $s$ is considered the evidence $E$ in Eq. \ref{eq:LR}, which results in:
\begin{equation}
    LR = \frac{Pr(s|H_p)}{Pr(s|H_d)}.
\label{eq:Score-based_LR}
\end{equation}

We propose to combine the facial descriptors of the available images of each person before computing the biometric score $s$.

In this work, we obtain this score as follows: let $X^r$ be the reference image and $X^t$ a trace image. A face recognition model $F$ is used to encode each image into compact vector representations $\mathbf{v}^r$ and $\mathbf{v}^t$ for the reference image and the trace image, respectively:

\begin{subequations}
    \begin{equation}
        \textbf{v}^r = F(X^r)
    \end{equation}
    \begin{equation}
        \textbf{v}^t = F(X^t).
    \end{equation}
\end{subequations}
    
For the face recognition system used in this work, the similarity score between the trace and the reference image is computed as the cosine similarity, given by 

\begin{equation}
s = \frac{\mathbf{v}^r.(\mathbf{v}^t)^T}{||\mathbf{v}^r||\,||\mathbf{v}^t||}.
\label{eq:score}
\end{equation}

In Equation \ref{eq:score}, $||\;||$ is the vector L2-norm and $\mathbf{v}^r.(\mathbf{v}^t)^T$ is the internal product between $\mathbf{v}^r$ and $\mathbf{v}^t$.

When multiple trace images are available, it is possible to compute multiple scores, raising the question of which score to derive the LR from. To address this problem, forensic experts can obtain a single score considering only the trace image with the highest quality. They can also aggregate the individual scores using either the maximum (\textbf{\textit{MaxScore}}) or the average (\textbf{\textit{AvgScore}}) of the individual scores. While this strategy allows obtaining a single LR value based on multiple pieces of evidence, a lot of information is disregarded in the estimation of the final score. For this reason, we propose to obtain a single score based on a linear combination of the visual descriptors of all available trace images:

\begin{equation}
s^* = \frac{\mathbf{v}^r.(\mathbf{v}^*)^T}{||\mathbf{v}^r||\,||\mathbf{v}^*||},
\end{equation}

where $\mathbf{v}^*$ is obtained from:

\begin{equation}
    \mathbf{v}^* = \sum_{i=1}^{N} w_i\mathbf{v}^{t_i}.
\label{eq:aggregation}
\end{equation}

In Equation \ref{eq:aggregation}, $w_i$ is the weight of the visual descriptor of the corresponding trace image, and $N$ is the number of trace images available.

Contrary to traditional score aggregation approaches \citep{score_level_fusion_encyclopedia}, our insight is to aggregate the trace images' descriptors to obtain a more robust representation of the person's identity. For this, we introduce different strategies to determine the weights of each trace image to compute the final visual descriptors.

\subsection{\textit{Ser-Fiq} Pooling}

Based on the assumption that the  face image's visual quality should guide the image's contribution to the final visual descriptor, we relied on a state-of-the-art face image quality estimator \citep{serfiq}. We used the normalized \textit{Ser-Fiq} quality score $s_i$ of each trace image as $w_i$:
\begin{equation}
w_i = \frac{s_i}{\sum_{j=1}^{N} s_j}.    
\end{equation}

\subsection{CS Pooling} 

We also considered the recently proposed face quality estimator \textit{Confusion Score} (CS) \citep{RUIFROK2022111201_confusion_scores} as a weighting mechanism for aggregation. In this strategy, the weight $w_i$ of a trace image with Confusion Score $cs_i$ is computed according to Eq. \ref{eq.cspool}:

\begin{equation}
\label{eq.cspool}
w_i = \frac{1 - cs_i}{\sum_{j=1}^{N} (1-cs_j)}, 
\end{equation}
since images with a better quality yield lower Confusion Scores.

\subsection{Average Pooling}

In this strategy, all the trace images are assigned the same weight, effectively reducing to a simple, unweighted average of each component of the visual descriptors.

\section{Data}
\label{section_data}

We selected datasets that represent two typical scenarios in forensic casework: surveillance and social media images.

\subsection{Surveillance Datasets}

In surveillance scenarios, subjects' images are captured without control of pose, illumination, expression, and other factors affecting facial recognition performance. Additionally, motion blur, compression artifacts, and low resolution of the face region are typical limitations present in this kind of data. On the other hand, reference images of a suspect are usually of excellent quality and captured under controlled conditions (e.g., driver's license or passport photo). Despite the multiple datasets devised to study face recognition in the wild, few datasets mimic the conditions of real surveillance scenarios \citep{raji2021face}.

Quis-Campi~\citep{quis_campi} and SCFace~\citep{scface} are the most representative datasets comprising data replicating the surveillance scenarios' degradation factors while providing high-quality reference images. For these reasons, we rely on these datasets in our experiments. 

The SCface dataset contains CCTV images of 130 subjects, captured at three different distances (\textit{far} - 4.2 m, \textit{medium} - 2.6 m, and \textit{close} - 1.0 m) from multiple cameras in the visible and infra-red spectrum. Additionally, it provides high-quality reference images captured in frontal pose and at varying degrees of lateral poses \citep{scface}. In our experiments, only the high-quality frontal images are considered references in the 1:1 comparisons. As for the CCTV images, which we use as traces, we only use images from the five cameras in the visible light spectrum.

The Quis-Campi dataset contains CCTV images of 320 subjects captured in an uncontrolled outdoor environment. In addition to variations in pose and distance, also present in the SCface dataset, surveillance images from the Quis-Campi dataset have significant variations in illumination, occlusion, and facial expression. Motion blur is also present in some images. Each subject has one frontal and two lateral profile reference images, with controlled illumination and neutral expression. Only frontal images are used as references in this work.

\subsubsection{Novel Verification Protocol for the Quis-Campi dataset}

To evaluate the proposed method in a more realistic surveillance scenario, we present a new verification protocol\footnote{Metadata for this new protocol will be available in the accepted version.} for the Quis-Campi dataset, based on the concept of \textit{encounters}. In this protocol, the surveillance images of each identity are grouped into sets of images captured during an encounter of the person of interest with the camera. For this purpose, we selected a threshold of two minutes as the criteria for separating the encounters of each person in the dataset. Each group of trace images of an encounter is compared to the corresponding reference image, according to the strategies described in Section \ref{section_method}. This protocol is representative of cases where images of a perpetrator are registered in the video, and no other surveillance images that can be safely attributed to the same perpetrator are available. Results using this protocol are referred to as \textit{Quis-campi encounters}.

\subsection{Social Media Datasets}

Trace images obtained from social media platforms are usually better than those obtained in surveillance scenarios. Nevertheless, these data still exhibit large
variations in pose, illumination, facial expression, and resolution. Moreover, traditional and digital makeup/beautification effects are also frequently present in these images. Two datasets were selected to evaluate our approach in this scenario: Adience \citep{adience} and Balanced Faces in the Wild (BFW) \citep{bfw}.

The Adience dataset was created to study age and gender recognition in data obtained in real-world imaging conditions. For this, 26,580 photos of 2,284 subjects were obtained from online image repositories. Images were acquired using smartphones and other mobile devices and presented significant variations in pose, lighting condition, facial expression, and image quality.

Considering that the number of images per identity is heavily imbalanced, we selected a subset of the Adience dataset, including identities with at least 11 images - one for reference and at least ten as traces. This selection resulted in a set of 14,143 images from 373 identities.

The BFW dataset contains 20,000 images of 800 individuals labeled for gender (female, male) and ethnicity (Asian, Black, Indian, White). The dataset is balanced, with 25 images per subject and 100 subjects in each demographic group.

\subsubsection{Definition of References for Adience and BFW Datasets}

The concept of the reference image is absent in social media datasets. 
Based on the assumption that in forensic scenarios, the reference images are typically acquired in more controlled scenarios, we select the image with the highest face quality as the reference image from each identity.

\begin{figure*}
     \centering
     \begin{subfigure}[b]{0.373\textwidth}
         \centering
         \includegraphics[width=\textwidth]{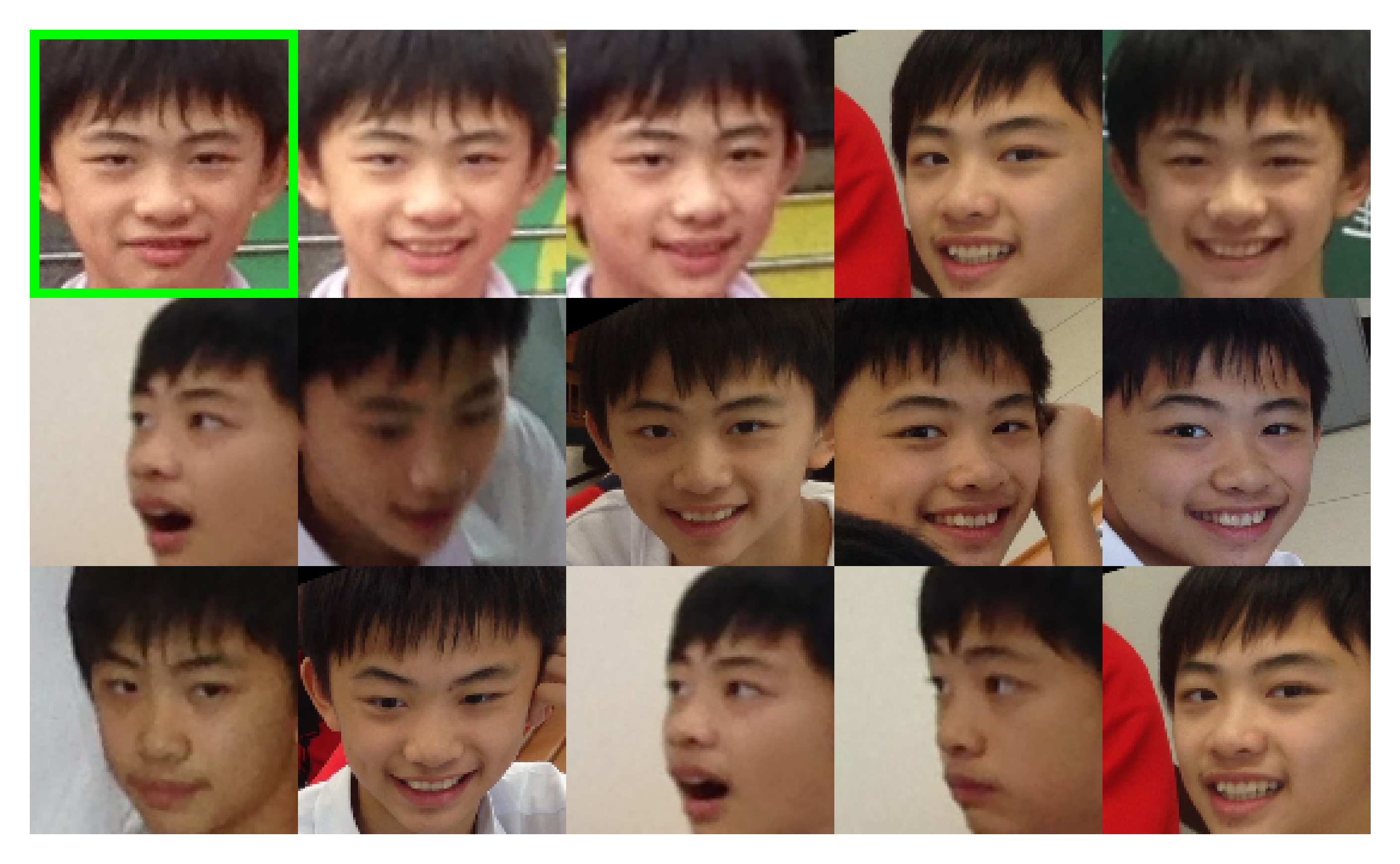}
     \end{subfigure}
     \hfill
     \begin{subfigure}[b]{0.587\textwidth}
         \centering
         \includegraphics[width=\textwidth]{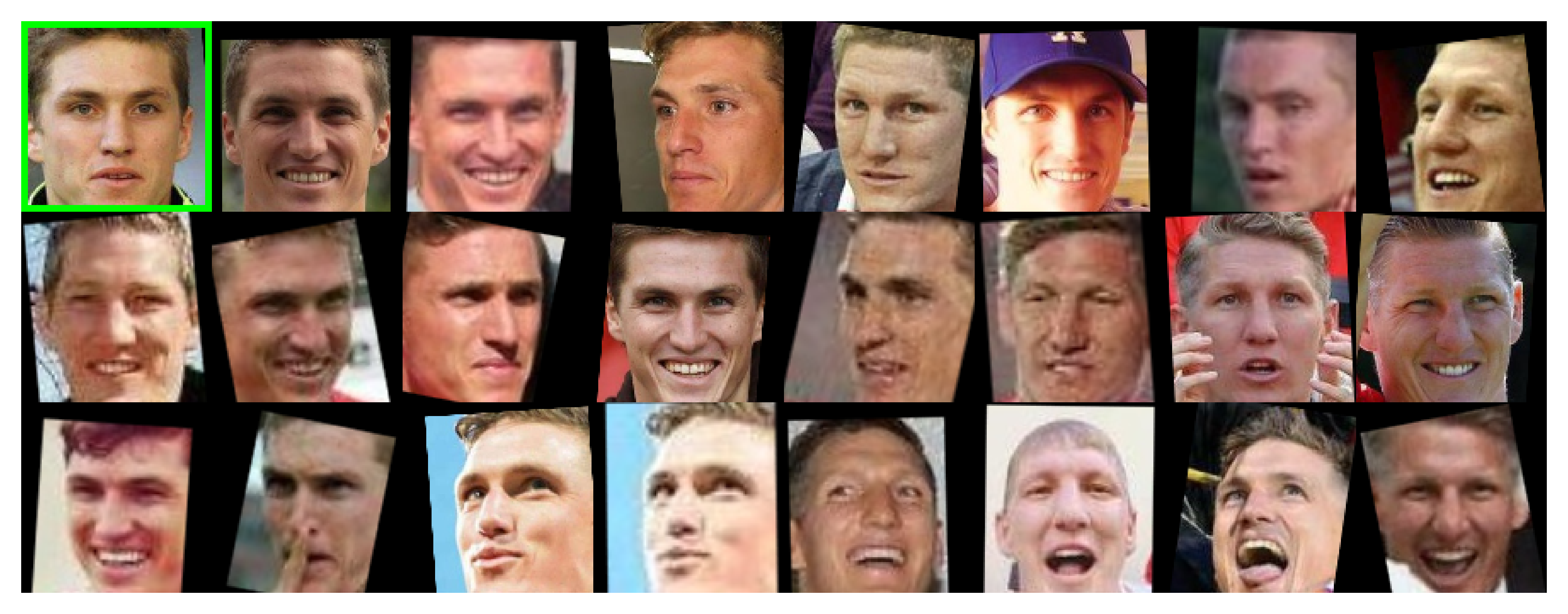}
     \end{subfigure}
     \begin{subfigure}[b]{0.373\textwidth}
         \centering
         \includegraphics[width=\textwidth]{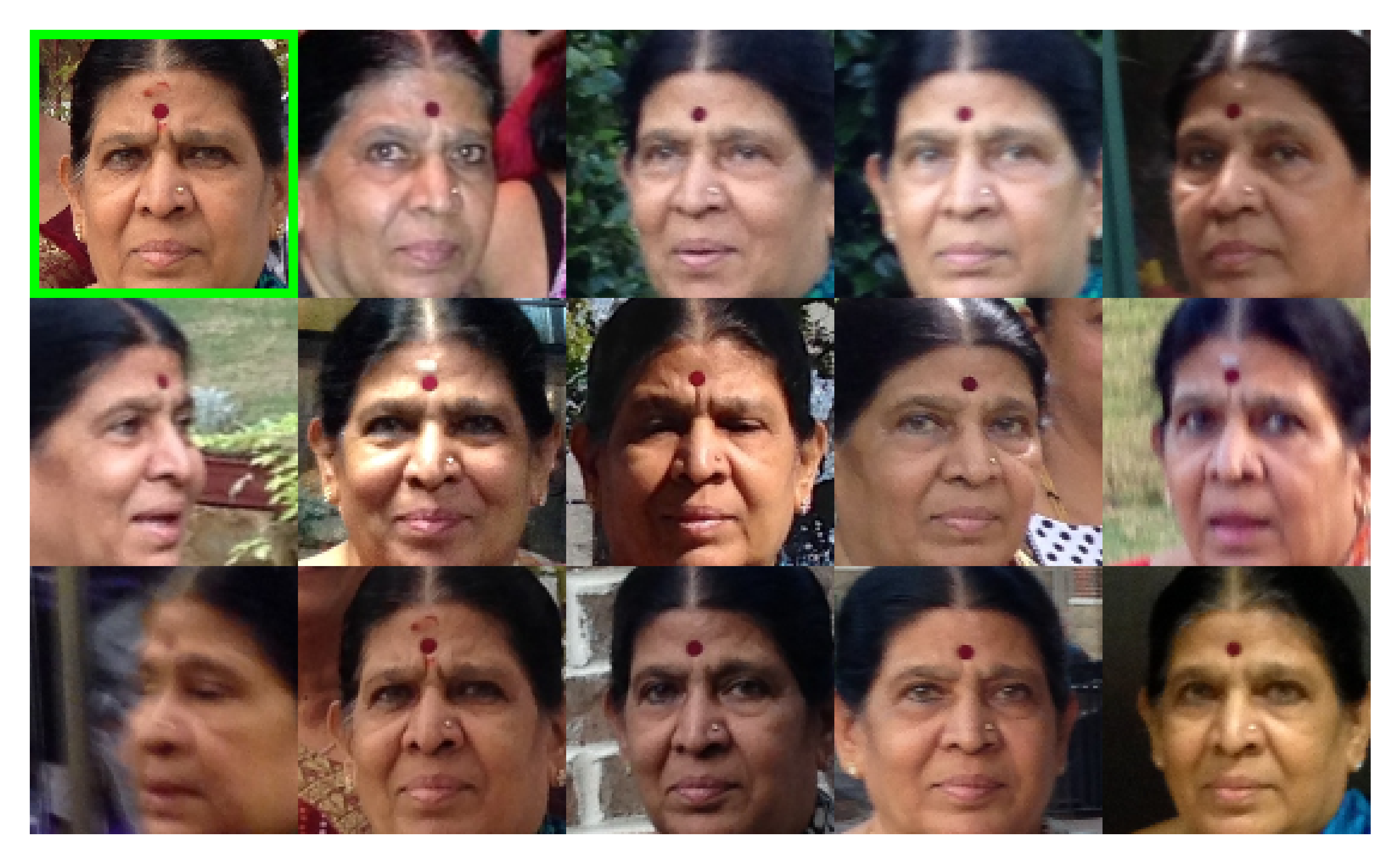}
         \caption{Adience}
     \end{subfigure}
     \hfill
     \begin{subfigure}[b]{0.587\textwidth}
         \centering
         \includegraphics[width=\textwidth]{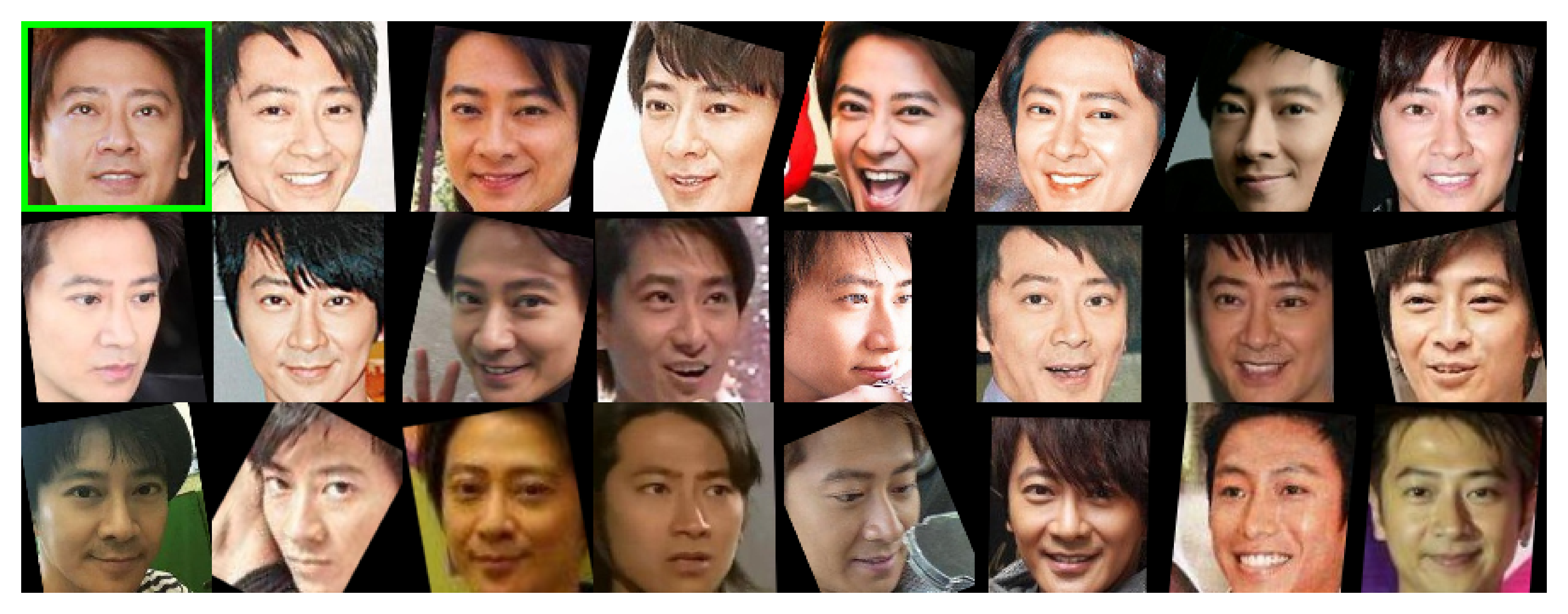}
         \caption{BFW}
     \end{subfigure}
        
        \caption{Examples of references selected for the Adience and BFW datasets. For each identity, the face at the top left (in green) is selected as a reference, and the others are used as traces.}
        \label{fig:refs_selection}
\end{figure*}

In particular, we rank the images according to their \textit{Ser-Fiq} and Confusion Scores and select the image with the best-combined ranking. Figure \ref{fig:refs_selection} depicts the selected references for two identities of each of the Adience and BFW datasets, illustrating that this strategy resulted in the selection of good-quality reference images.

\subsubsection{Identity Errors in Adience and BFW Datasets}

\begin{figure*}[h]
\label{fig:bimodalgenuinescores}
     \centering
     \begin{subfigure}[b]{0.48\textwidth}
         \centering
         \includegraphics[width=\textwidth]{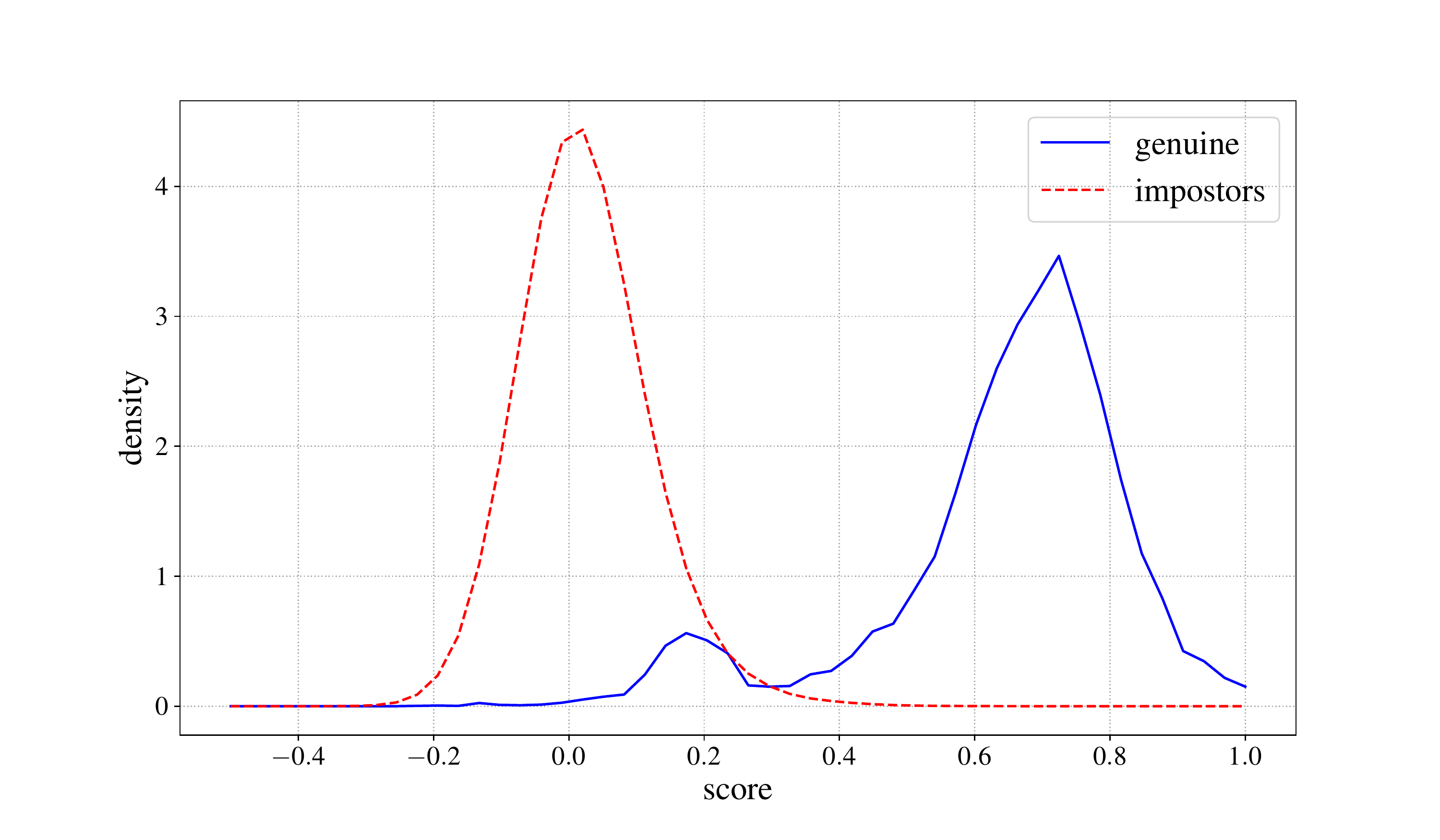}
         \caption{Adience}
         \label{fig:bimodal_genuine_scores_adience}
     \end{subfigure}
     \begin{subfigure}[b]{0.48\textwidth}
         \centering
         \includegraphics[width=\textwidth]{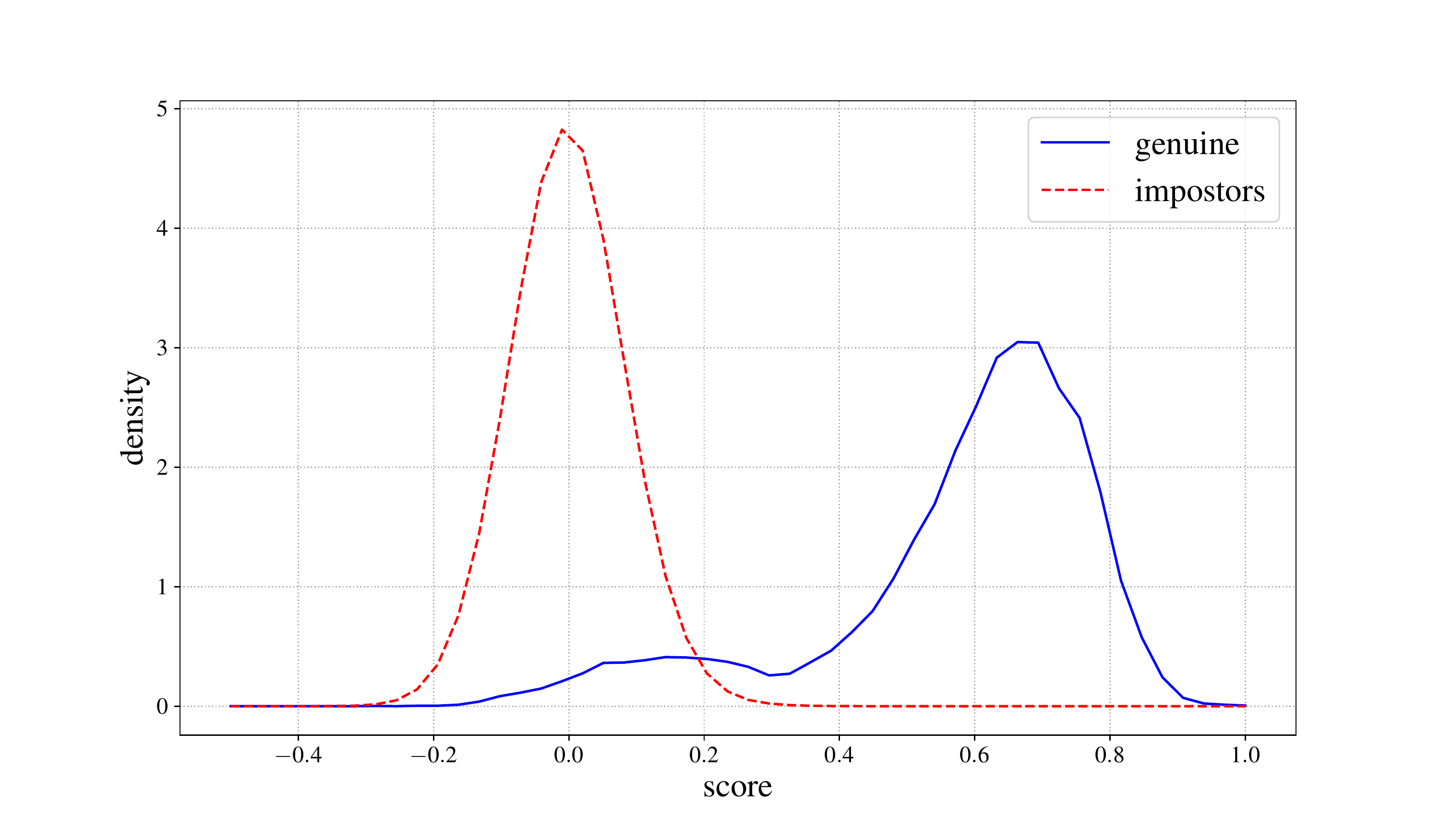}
         \caption{BFW}
         \label{fig:bimodal_genuine_scores_bfw}
     \end{subfigure}
     \begin{subfigure}[b]{0.48\textwidth}
         \centering
         \includegraphics[width=\textwidth]{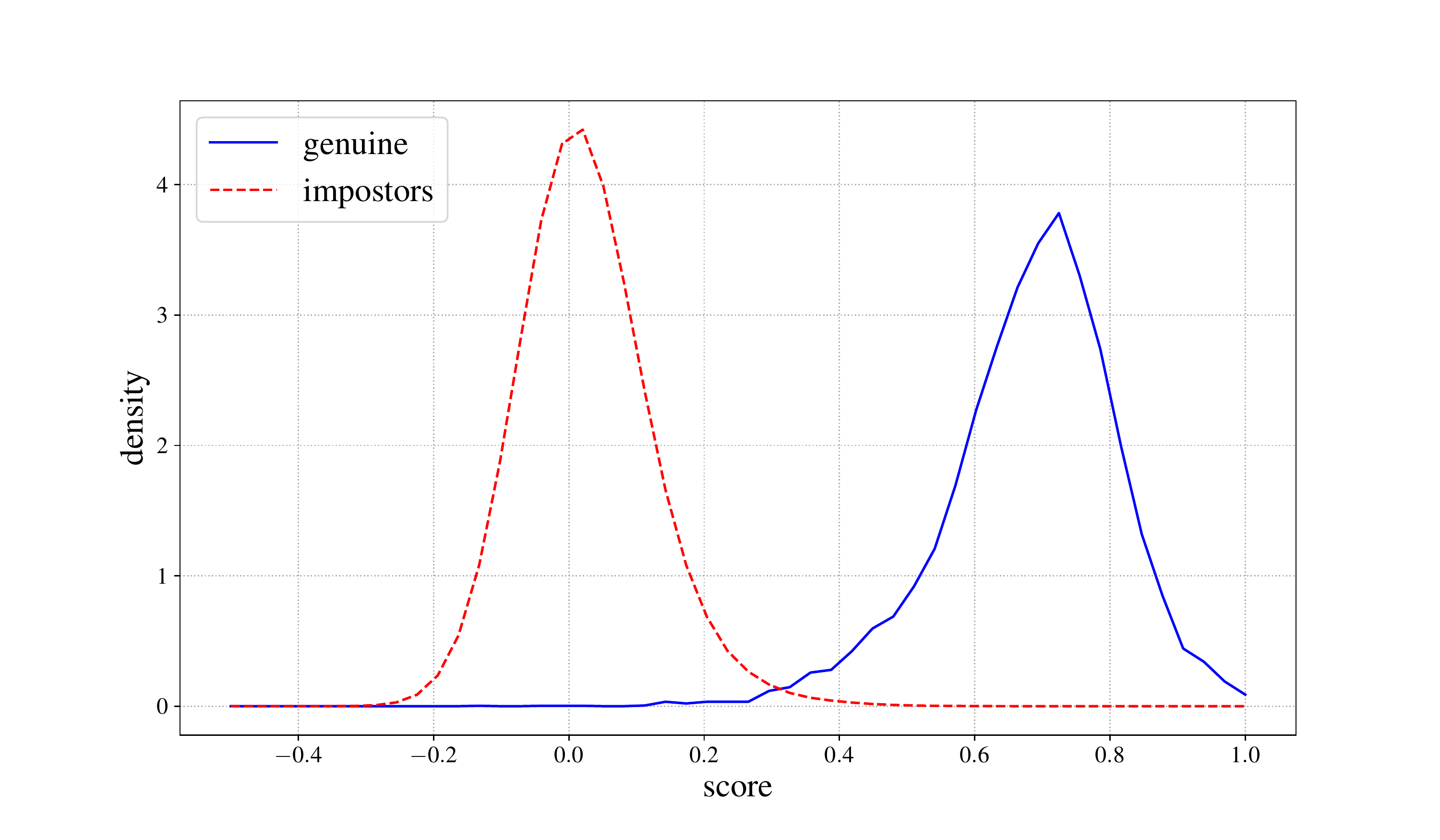}
         \caption{Adience clean}
         \label{fig:genuine_scores_cleaned_adience}
     \end{subfigure}
     \begin{subfigure}[b]{0.48\textwidth}
         \centering
         \includegraphics[width=\textwidth]{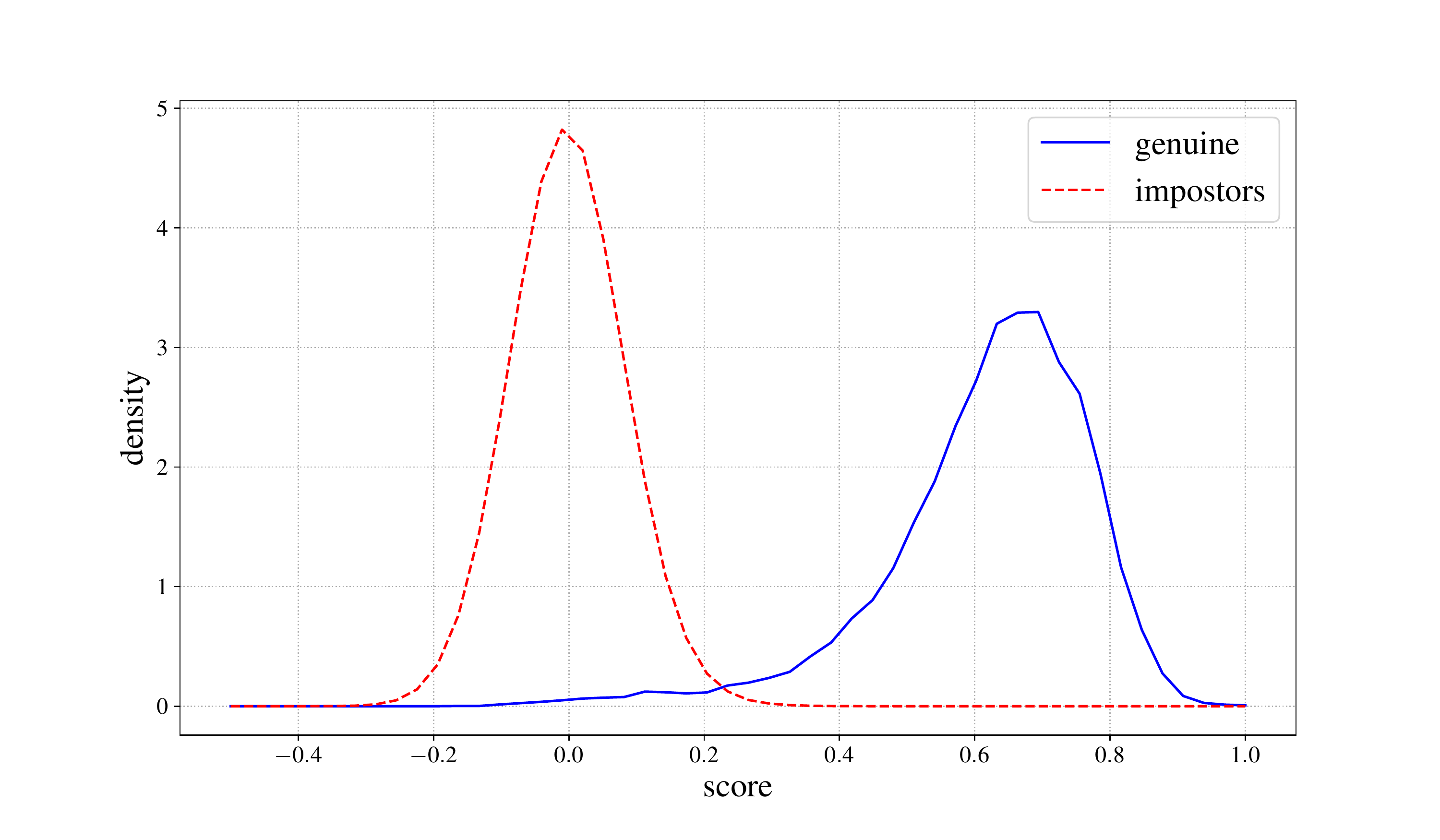}
         \caption{BFW clean}
         \label{fig:genuine_scores_cleaned_bfw}
     \end{subfigure}
        
    \caption{Bi-modal behavior of genuine scores distributions for the Adience (a) and BFW (b) datasets, suggestive of identity labeling errors. After cleaning, the genuine distributions no longer exhibit this bi-modal behavior (c, d).}
    \label{fig:bimodal_genuine_scores}
\end{figure*}

During our preliminary experiments on the Adience and BFW datasets, we observed an atypical bi-modal distribution of the genuine scores (Figures \ref{fig:bimodal_genuine_scores_adience} and \ref{fig:bimodal_genuine_scores_bfw}). This unexpected behavior raised suspicion that errors in the identity labels might be present in these two datasets.

\begin{figure*}
    \centering
    \begin{subfigure}[b]{0.33\textwidth}
     \centering
     \begin{subfigure}[b]{0.38\textwidth}
         \centering
         \includegraphics[width=\textwidth]{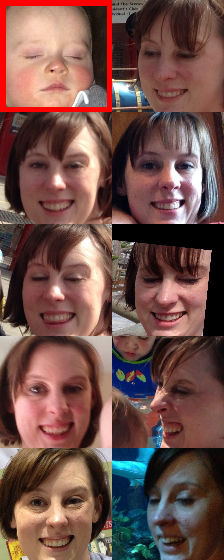}
     \end{subfigure}
     \hfill
     \begin{subfigure}[b]{0.57\textwidth}
         \centering
         \includegraphics[width=\textwidth]{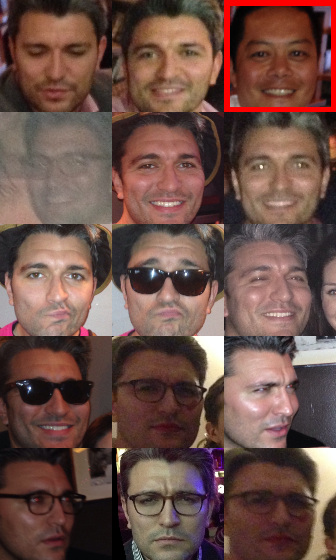}
     \end{subfigure}
    \caption{Adience}
    \end{subfigure}
    \hfill
    \begin{subfigure}[b]{0.64\textwidth}
     \begin{subfigure}[b]{0.49\textwidth}
         \centering
         \includegraphics[width=\textwidth]{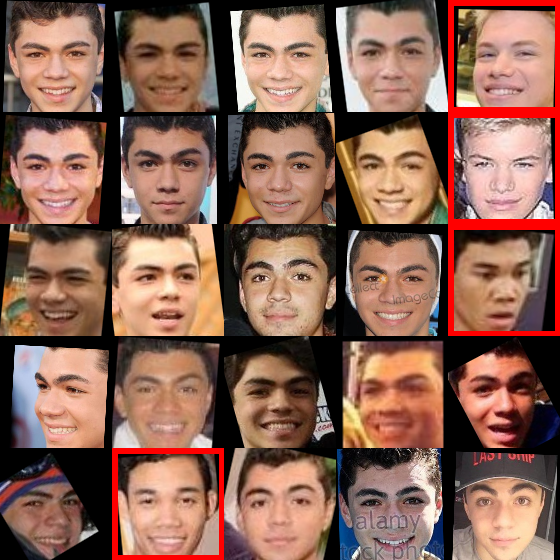}
     \end{subfigure}
     \hfill
     \begin{subfigure}[b]{0.49\textwidth}
         \centering
         \includegraphics[width=\textwidth]{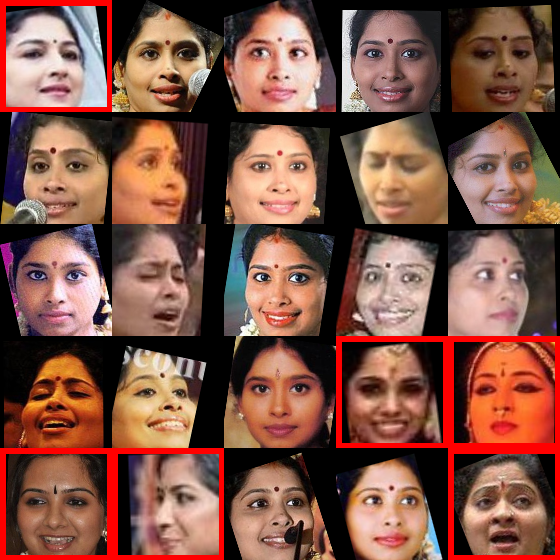}
     \end{subfigure}
     \caption{BFW}
    \end{subfigure}
        \caption{Examples of identity labeling errors (red boxes) in the Adience and BFW datasets.}
        \label{fig:errors_adience_bfw}
\end{figure*}

A manual review of the images most frequently involved in low genuine scores confirmed that many identity labels were incorrect in both datasets. Figure \ref{fig:errors_adience_bfw} shows some examples of these errors.

We adopt a strategy to clean the datasets automatically to mitigate the effects of such errors. We rely on the approach proposed in \citep{community_cleaning} that allows the re-assignment of the identity label for images initially deemed incorrectly labeled, minimizing the number of images discarded from the original datasets. Additionally, we manually identified and removed 841 duplicated (same hash) images in the Adience dataset.\footnote{The list of duplicated images is available at [redacted for submission]} The cleaned versions of the Adience and BFW datasets, hereafter referred to as \textit{Adience clean} and \textit{BFW clean}, are composed of 13,160 images from 355 identities, 19,131 images from 800 identities, respectively.

To assess the effectiveness of the cleaning process, we observe the differences between the distribution of the genuine and impostors scores before and after cleaning the datasets. The distributions of genuine scores of both cleaned datasets present a typical uni-modal distribution (Figures \ref{fig:genuine_scores_cleaned_adience} and \ref{fig:genuine_scores_cleaned_bfw}), indicating that the automated cleaning process succeeded in determining the mislabeled images.

\begin{figure*}[h]
     \centering
     \begin{subfigure}[b]{0.48\textwidth}
         \centering
         \includegraphics[width=\textwidth]{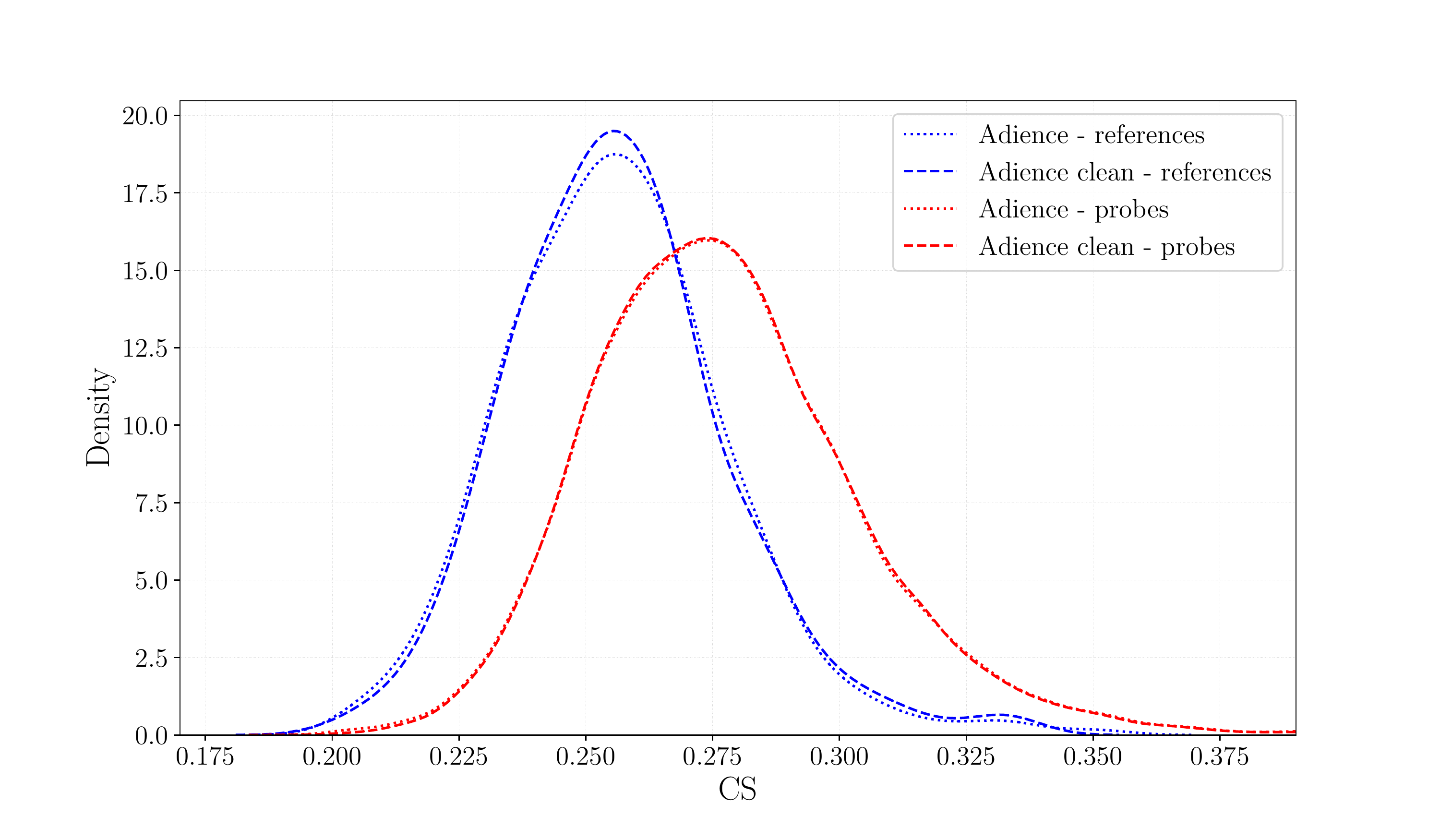}
         \caption{Adience}
     \end{subfigure}
     \begin{subfigure}[b]{0.48\textwidth}
         \centering
         \includegraphics[width=\textwidth]{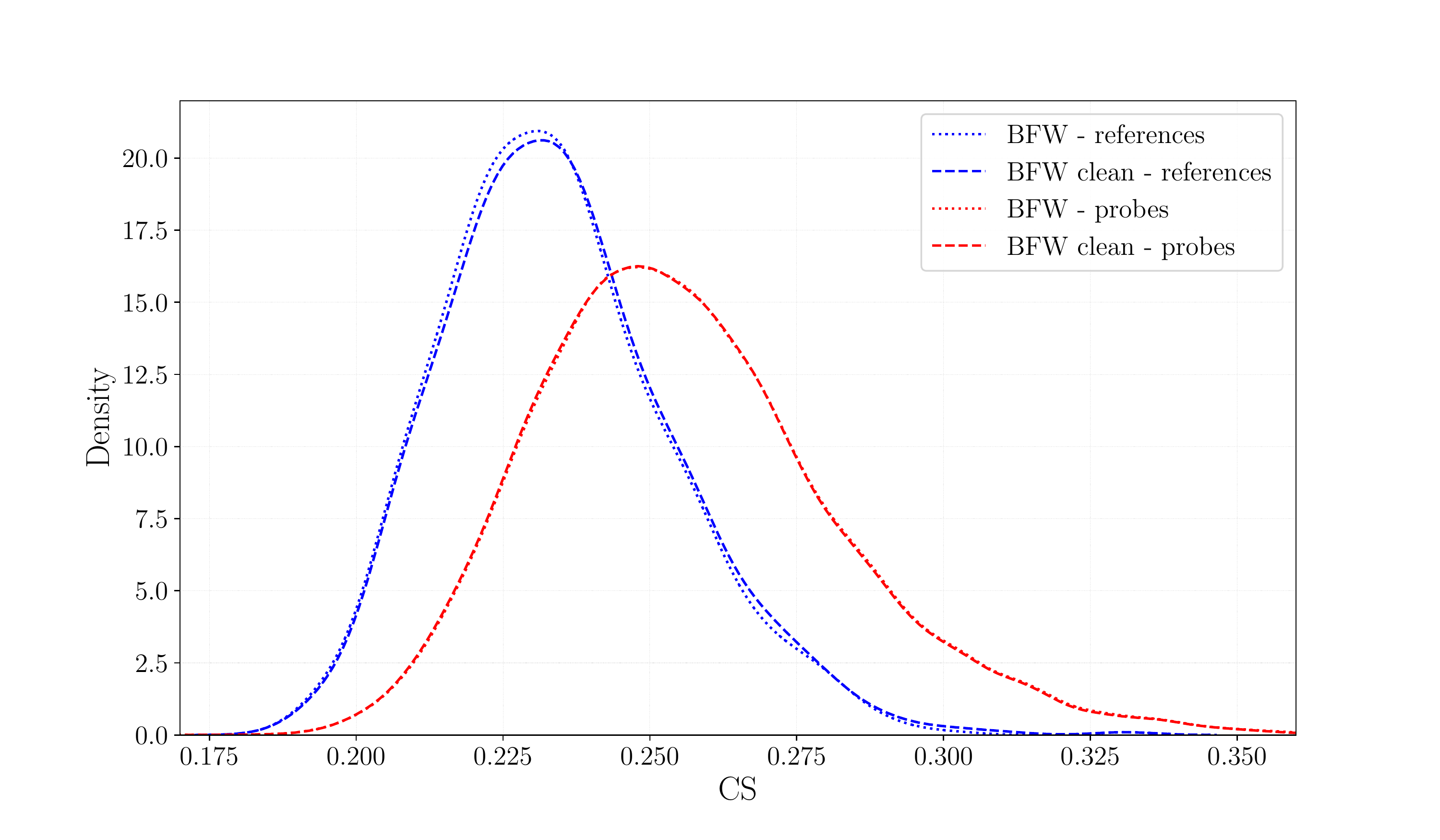}
         \caption{BFW}
     \end{subfigure}
    \caption{Distributions of Confusion Scores for the references and probes from the BFW and Adience datasets, before and after cleaning.}
    \label{fig:cs_before_after_cleaning}
\end{figure*}

To evaluate if the cleaning procedure had changed the difficulty for face recognition of the datasets, we investigated the differences in the distribution of Confusion Scores of the reference and probe images before and after cleaning. As depicted in Figure \ref{fig:cs_before_after_cleaning}, the distributions of the quality scores (CS) before and after the cleaning process are highly similar, suggesting that the cleaning procedure did not change the intrinsic difficulty of the datasets.

\section{Experiments}
\label{sec_experiments}

We focus our experiments on 1:1 comparisons between a reference image and a set of trace images. In forensic settings, these sets of trace images may originate from a surveillance video, with multiple frames depicting the person of interest, or from a set of images collected from social media profiles.

As a baseline, we rely on the independent comparison between a reference image and all the trace images from the dataset. This baseline is representative of common practices in forensic laboratories, where a single trace image, usually selected on the basis of its quality for comparison, is evaluated independently against the reference without any form of aggregation.

\begin{figure}[h]
     \centering
         \includegraphics[width=\columnwidth]{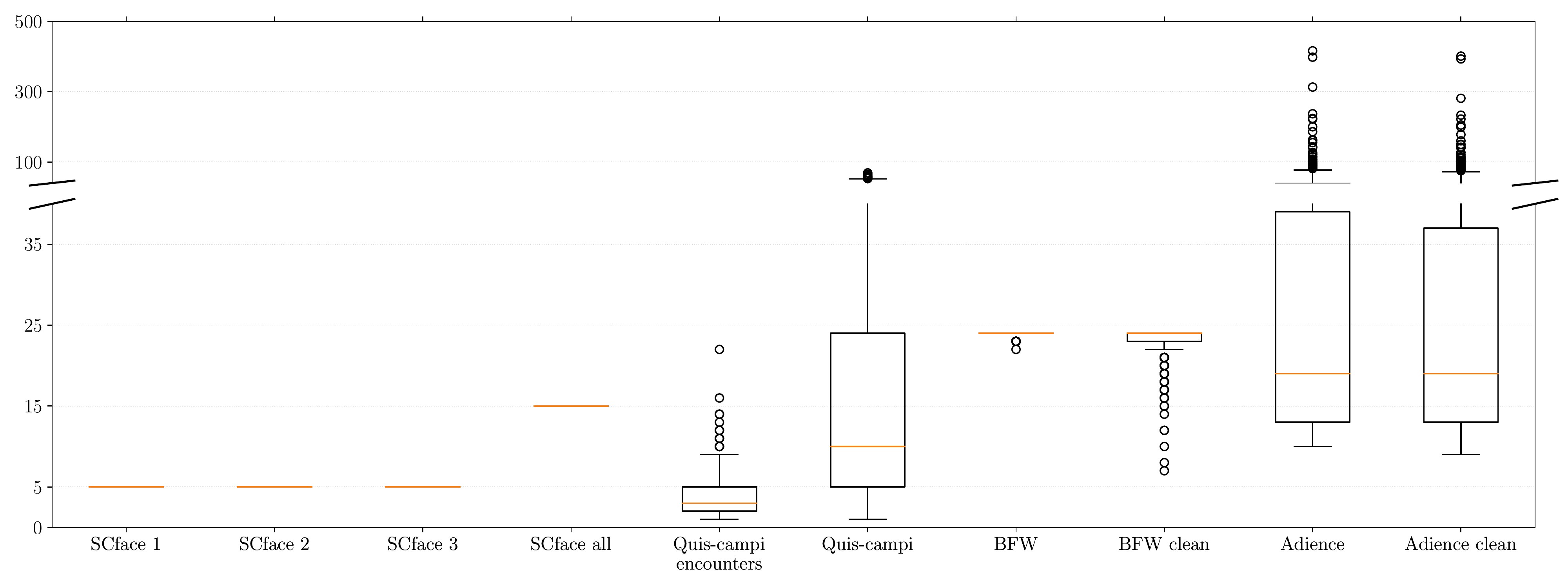}
         \caption{Distribution of the number of embeddings aggregated per identity in each dataset.}
         \label{fig:probes_per_id_per_dataset}
\end{figure}

We evaluate different strategies to integrate the information available in multiple images of the trace sets, as described in Section \ref{section_method}. We also evaluate the \textit{MaxScore} and \textit{AvgScore} strategies. Even though these are not based on embedding aggregation, we assume they are interesting due to their simplicity and applicability to forensic casework. The distribution of the number of embeddings that are aggregated per identity in each dataset is shown in Figure \ref{fig:probes_per_id_per_dataset}.

For each aggregation strategy, we calibrate the scores into LRs using a regularized logistic regression model, described in section \ref{subsec:score_to_lr_model}. For the SCface dataset, with a smaller number of identities, we calculate validation LRs using \textit{leave-one-identity-out} and \textit{leave-two-identities-out} cross-validation strategies. Due to computational limitations, we adopt a 100-fold cross-validation strategy for the other datasets to compute LRs. We evaluate the performance of the resulting score-based likelihood ratio systems using the log-likelihood ratio cost (\Cllr) \citep{Brummer_CLLR} and Tippett plots \citep{Tippett}.

\subsection{Face Recognition Model}
Since our focus is on aggregation strategies, we conduct all the experiments using a single face recognition model. In particular, we relied on SCRFD~\citep{scrfd} for face detection, and we used an affine transformation to align and crop the facial region into 112x112 images. For recognition, a ResNet-101 was trained on the MS1MV2 dataset \citep{arcface}, using the Arcface loss \citep{arcface}, achieving an accuracy of 99.83\% on the LFW dataset \citep{LFW}, which is on par with performance reported by state-of-the-art face recognition models.

\subsection{Score-to-LR Model}
\label{subsec:score_to_lr_model}

We use a regularized logistic regression model to calibrate a score into an LR, implemented in the open-source LIR Python package\footnote{Available in \url{https://github.com/netherlandsforensicinstitute/lir}}. Logistic regression models have traditionally been used in forensic speaker comparison \citep{emulating_dna,lr_fusion_nist99, tutorial_log_regression_morrison}. It does not assume a specific distribution of the training scores and is less susceptible to sampling variability \citep{ali_2014, Molder_Leitet_2020_score_to_LR}. Regularization\footnote{Since we were not interested in comparing the performance of different approaches of Score-to-LR calibration, we used the same regularization parameter of 1 to all experiments.} is used to induce shrinkage of the LR values, as in \citep{morrison_shrunk_2018}.

\section{Results and Discussion}
\label{sec_results}

\begin{figure*}
     \centering
     \begin{subfigure}[b]{0.48\textwidth}
         \centering
         \includegraphics[width=\textwidth]{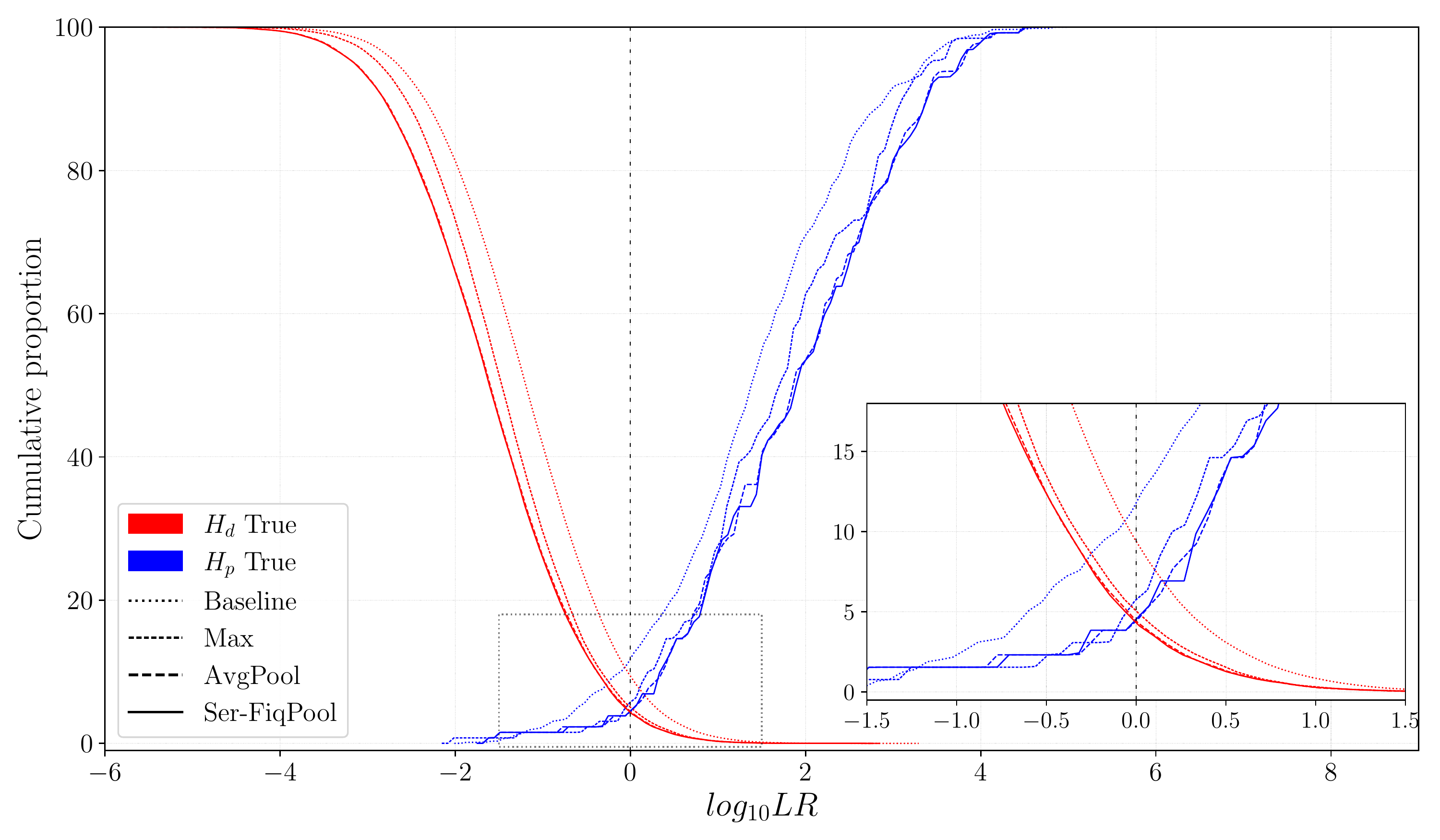}
         \caption{SCface 1}
     \end{subfigure}
     \hfill
     \begin{subfigure}[b]{0.48\textwidth}
         \centering
         \includegraphics[width=\textwidth]{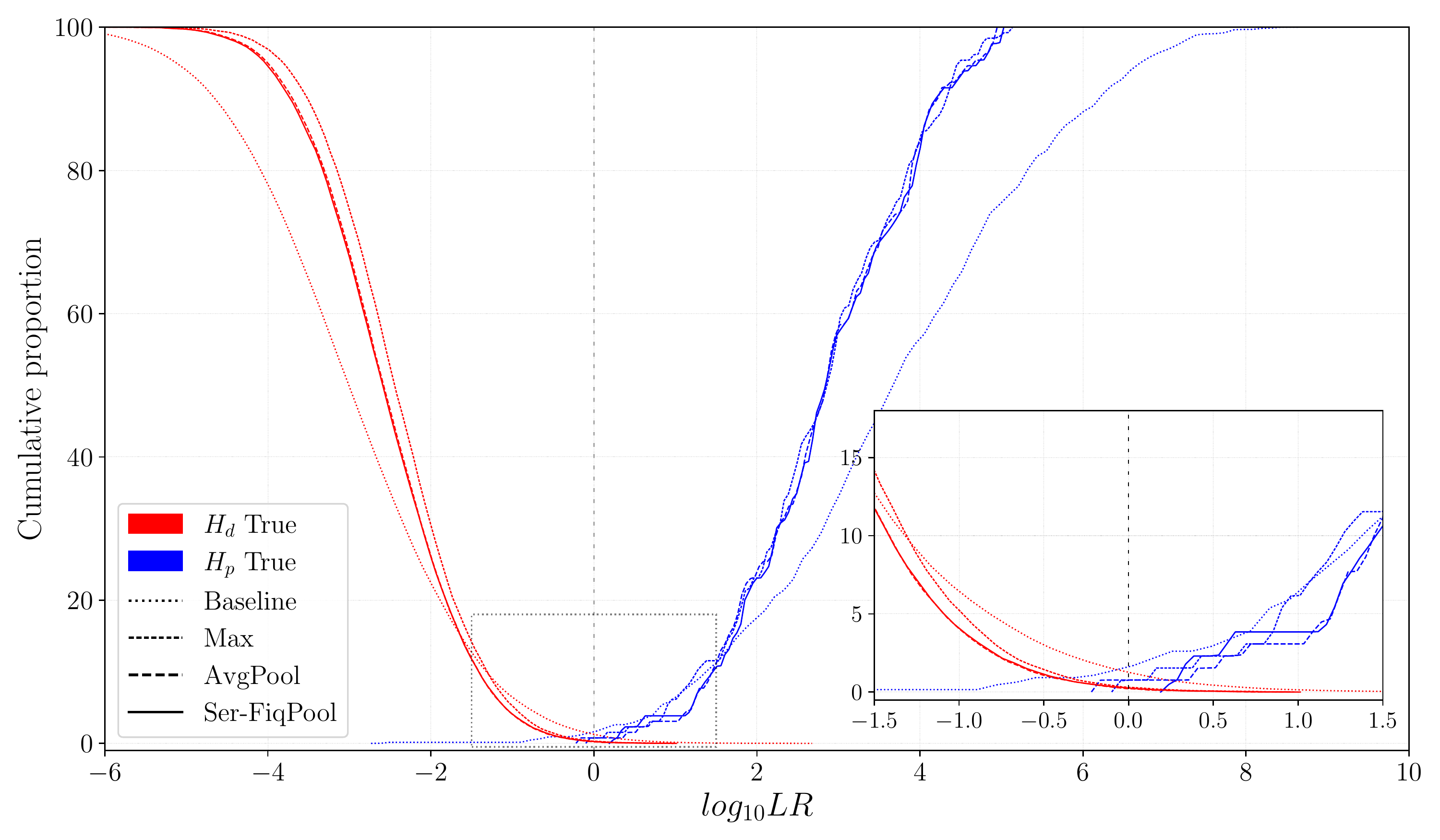}
         \caption{SCface 2}
     \end{subfigure}
     \begin{subfigure}[b]{0.48\textwidth}
         \centering
         \includegraphics[width=\textwidth]{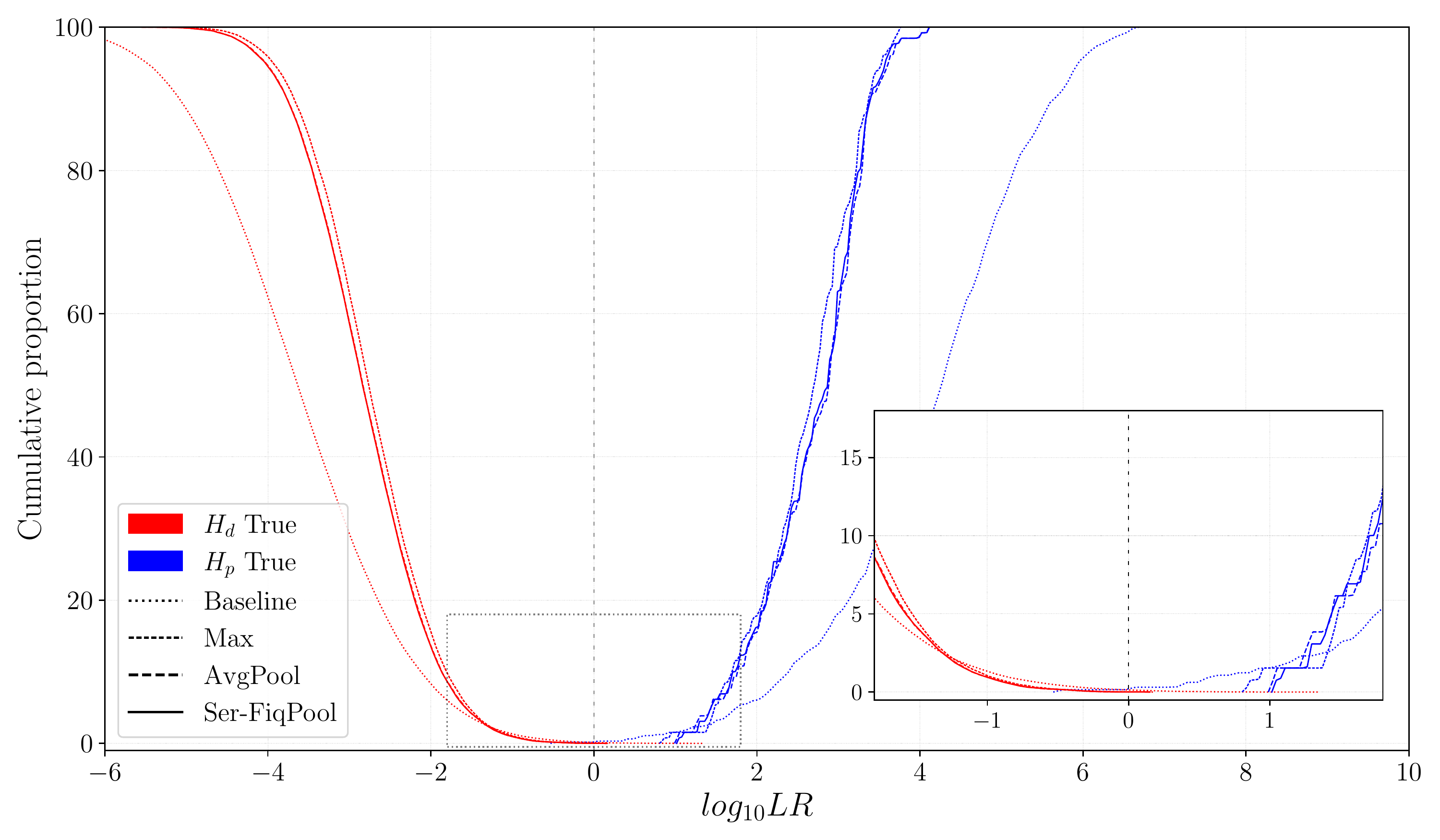}
         \caption{SCface 3}
     \end{subfigure}
     \hfill
     \begin{subfigure}[b]{0.48\textwidth}
         \centering
         \includegraphics[width=\textwidth]{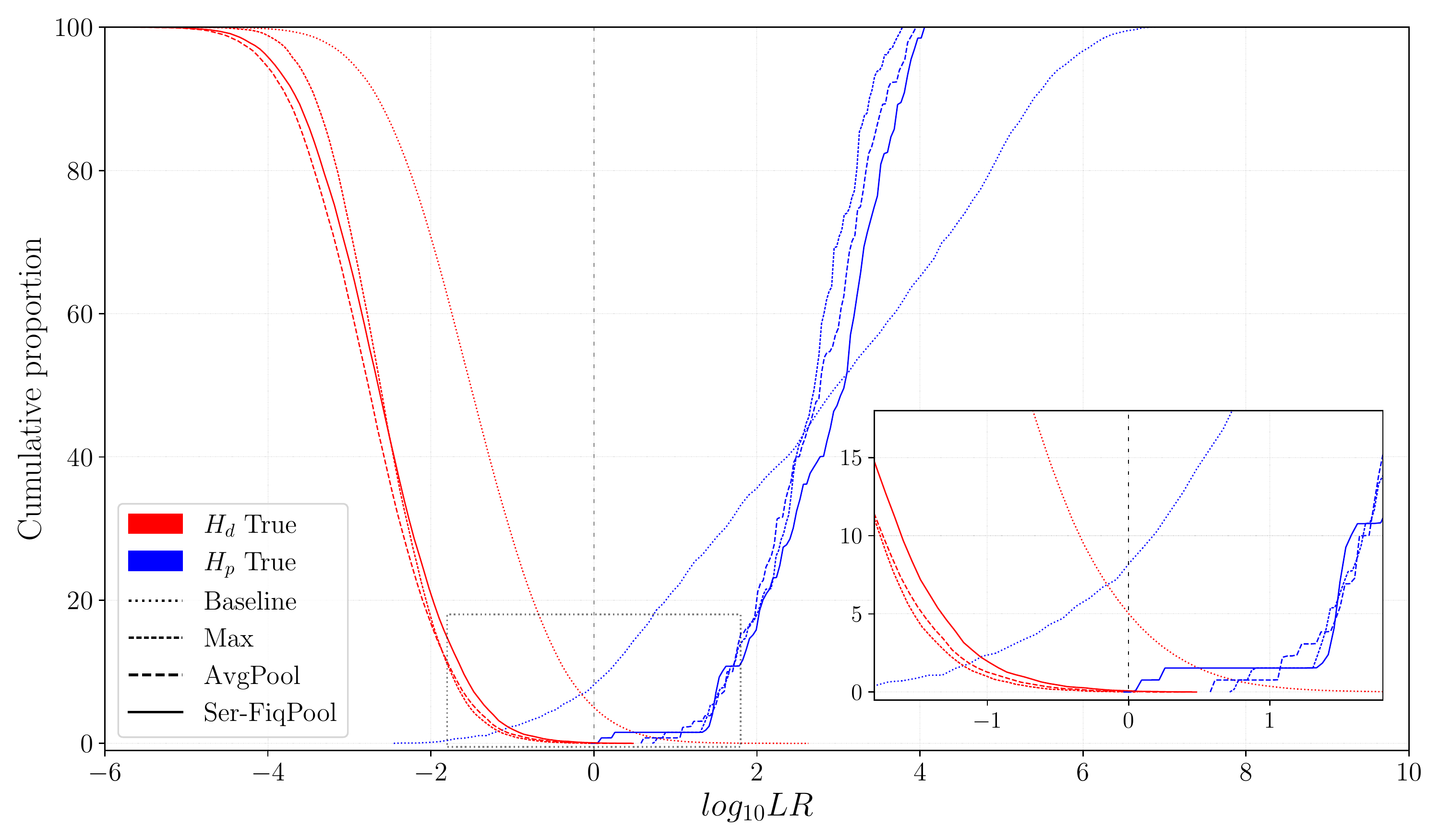}
         \caption{SCface all}
     \end{subfigure}
     \hfill
     \begin{subfigure}[b]{0.48\textwidth}
         \centering
         \includegraphics[width=\textwidth]{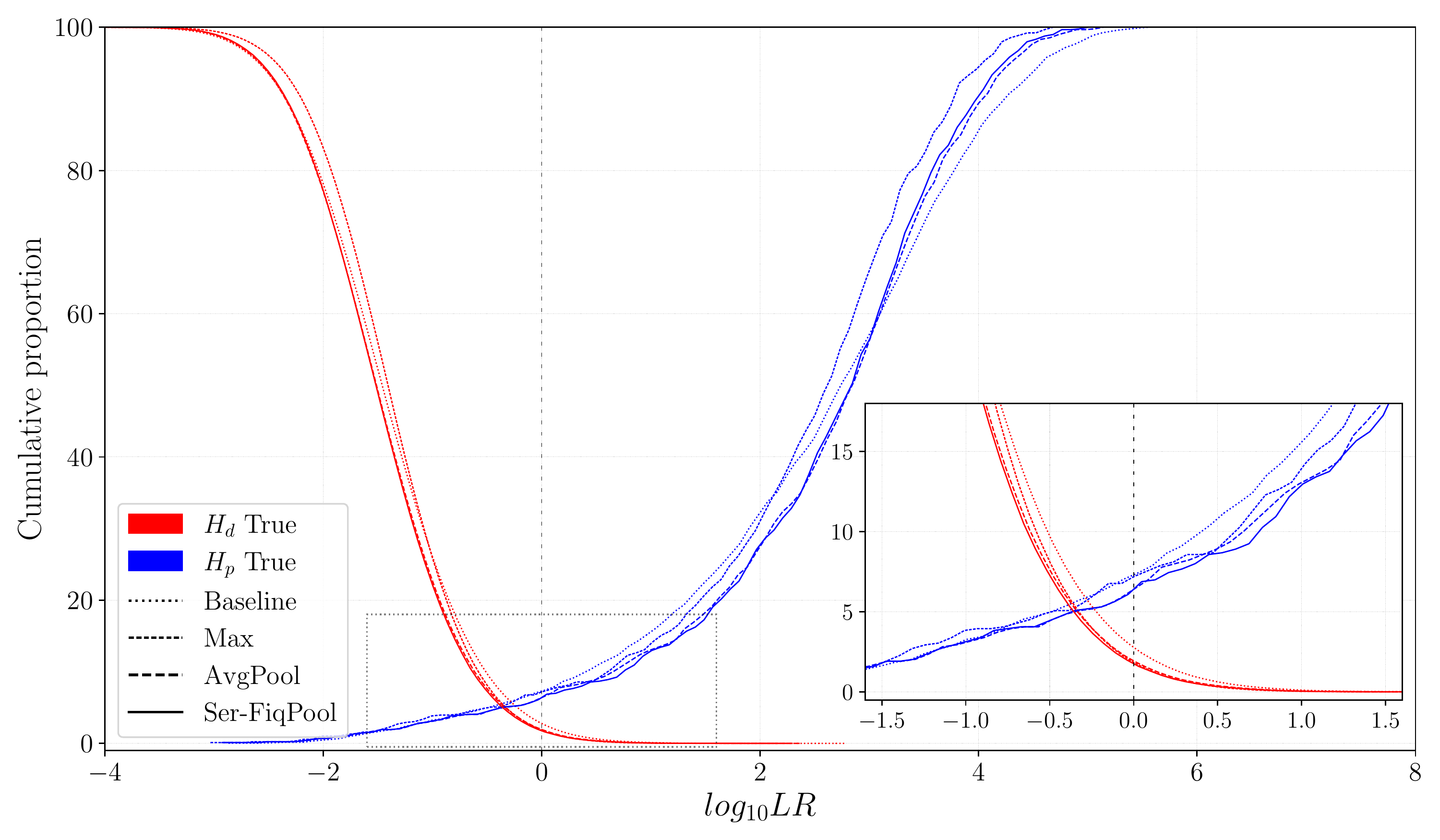}
         \caption{Quis-Campi encounters}
     \end{subfigure}
     \begin{subfigure}[b]{0.48\textwidth}
         \centering
         \includegraphics[width=\textwidth]{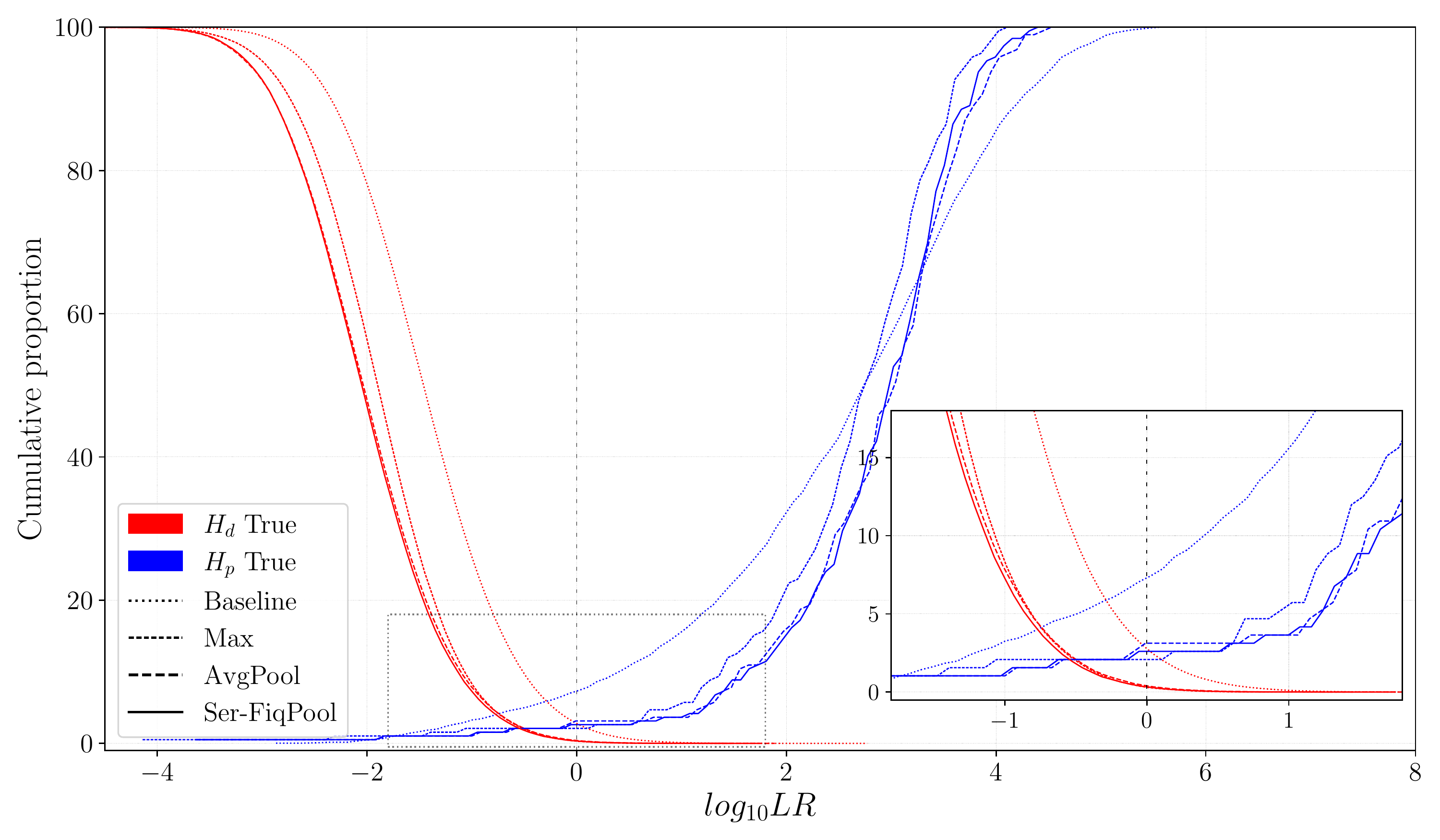}
         \caption{Quis-Campi}
     \end{subfigure}
        
        \caption{Tippett plots for the datasets of the surveillance scenario. The box on each plot shows details around $log_{10}\,LR=0$. Tippett plots for some strategies are omitted for clarity in the figure. Individual plots for every strategy are provided in the supplementary material file.}
        \label{fig:tippet_surveillance}
\end{figure*}

Results for the datasets of the surveillance scenario are shown in Table \ref{tab:cllr_surv} and Figure \ref{fig:tippet_surveillance}. We first observe the vast improvements in \Cllr compared to Mandasari et al. \citep{Mandasari_IETB2014} on the SCface dataset \citep{scface}. This improvement is mainly attributed to the discriminating power of the facial recognition module since even our baseline approach offered substantially better results. Regarding the embedding aggregation approaches, we observe that quality-based weighting (CSPool and Ser-FiqPool) showed the best performance overall but only marginally better than the other approaches. We also note that larger improvements in \Cllr occurred for the surveillance scenario datasets with more embeddings to be aggregated (SCface all and Quis-Campi). We also note a substantial gain for images with lower resolutions (SCface 1 and SCface 2).

\begin{table}[h]
\centering
\caption{\Cllr for the surveillance scenario}\label{tab:cllr_surv}
\resizebox{\columnwidth}{!}{%
\begin{tabular}{|c| c| c| c| c| c| c|}
\hline
 & SCface 1 & SCface 2 & SCface 3 & SCface & Quis-Campi & Quis-Campi \\
&           &          &          & all    & encounters &          \\
\toprule
\cite{Mandasari_IETB2014} &&&&&& \\
Raw scores & 0.659 & 0.313 & 0.378 & 0.503 & - & - \\
ZT-norm scores & 0.664 & 0.243 & 0.287 & 0.419 & - & - \\ \hline
Baseline & 0.368 & 0.060 & 0.011 & 0.249 & 0.226 & 0.226 \\ 
AvgScore & 0.221 & 0.037 & 0.013 & 0.023 & 0.209 & 0.105\\ 
MaxScore & 0.234 & 0.035 & 0.011 & \textbf{0.012} & 0.222 & 0.115\\ \hline
AvgPool & 0.212 & 0.029 & \textbf{0.010} & 0.013 & 0.202 & 0.098\\ 
CSPool & 0.210 & 0.029 & \textbf{0.010} & \textbf{0.012} & 0.201 & 0.098 \\ 
Ser-FiqPool & \textbf{0.209} & \textbf{0.028} & \textbf{0.010} & 0.018 & \textbf{0.198} & \textbf{0.095}\\ \hline

\end{tabular}
}
\end{table}

\begin{figure*}
     \centering
     \begin{subfigure}[b]{0.48\textwidth}
         \centering
         \includegraphics[width=\textwidth]{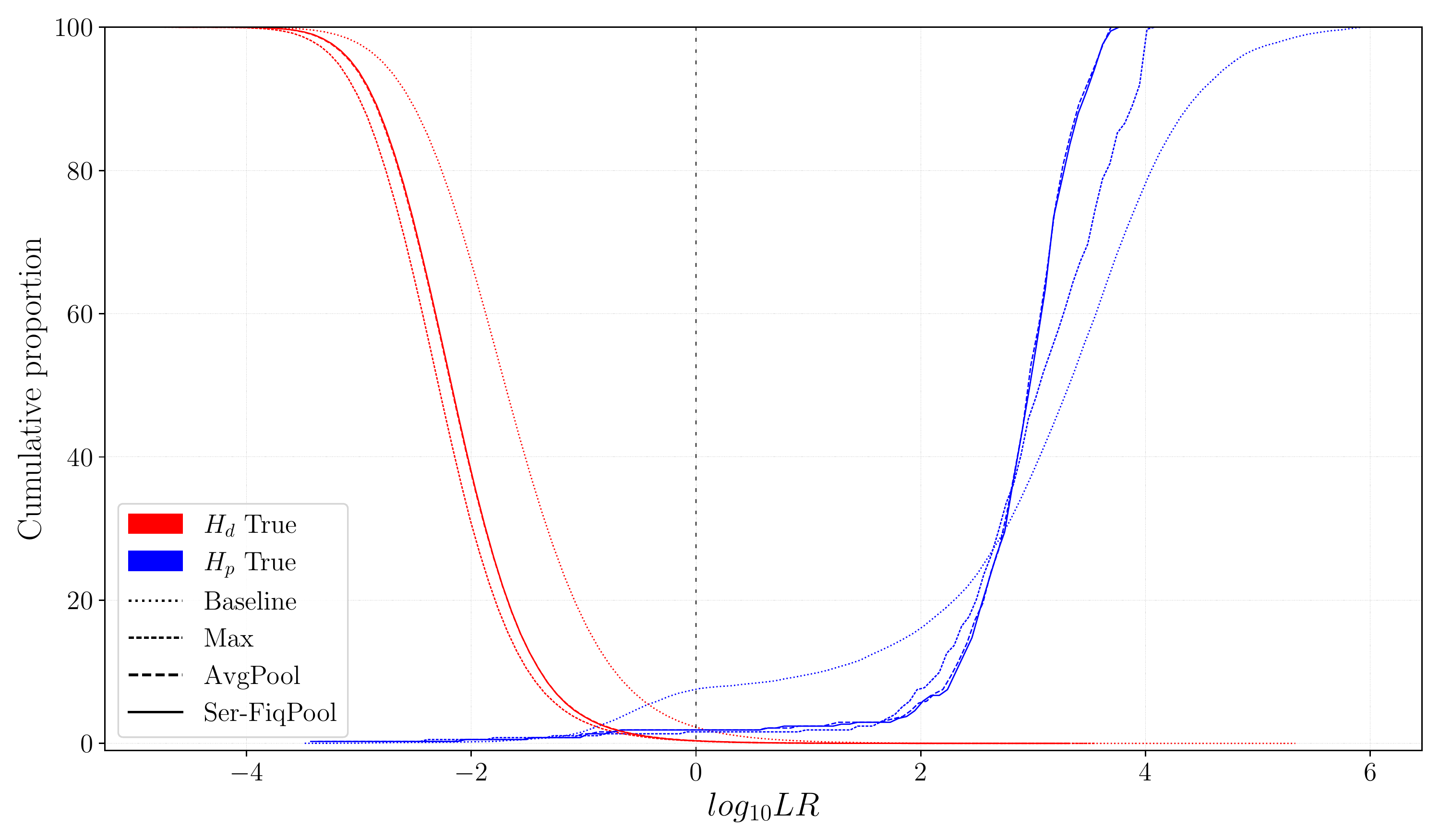}
         \caption{Adience}
     \end{subfigure}
     \hfill
     \begin{subfigure}[b]{0.48\textwidth}
         \centering
         \includegraphics[width=\textwidth]{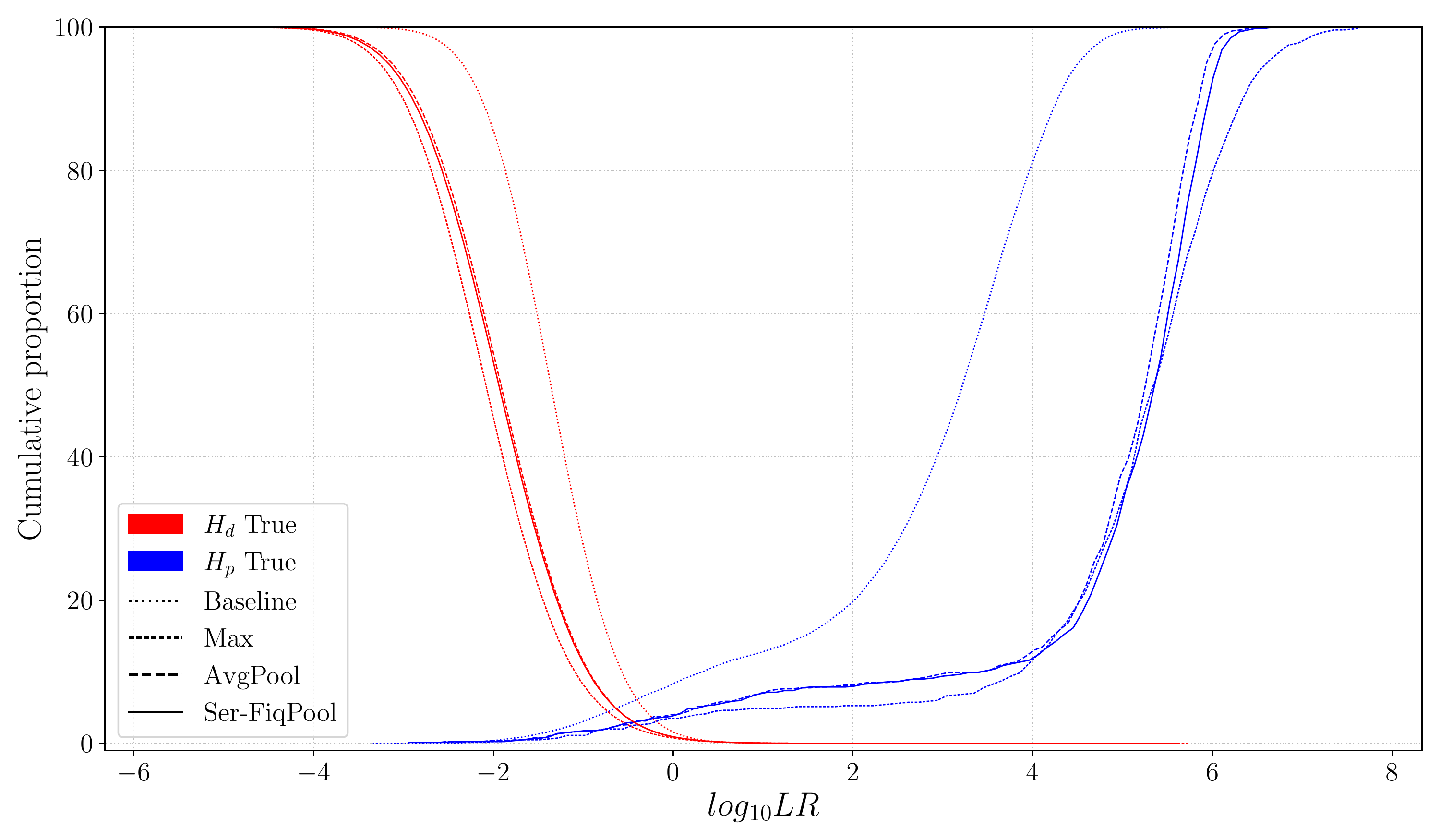}
         \caption{BFW}
     \end{subfigure}
     \begin{subfigure}[b]{0.48\textwidth}
         \centering
         \includegraphics[width=\textwidth]{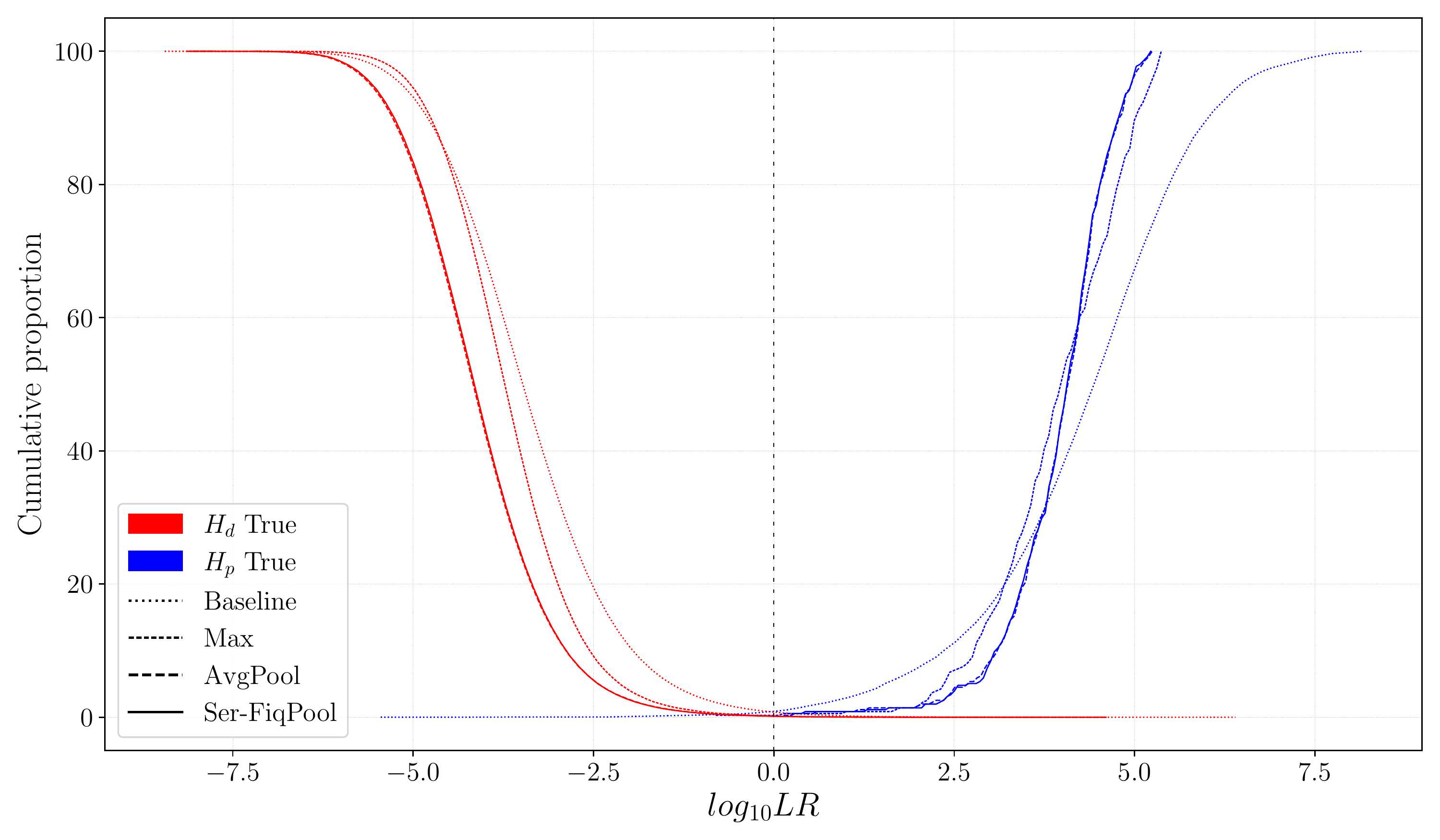}
         \caption{Adience cleaned}
     \end{subfigure}
     \hfill
     \begin{subfigure}[b]{0.48\textwidth}
         \centering
         \includegraphics[width=\textwidth]{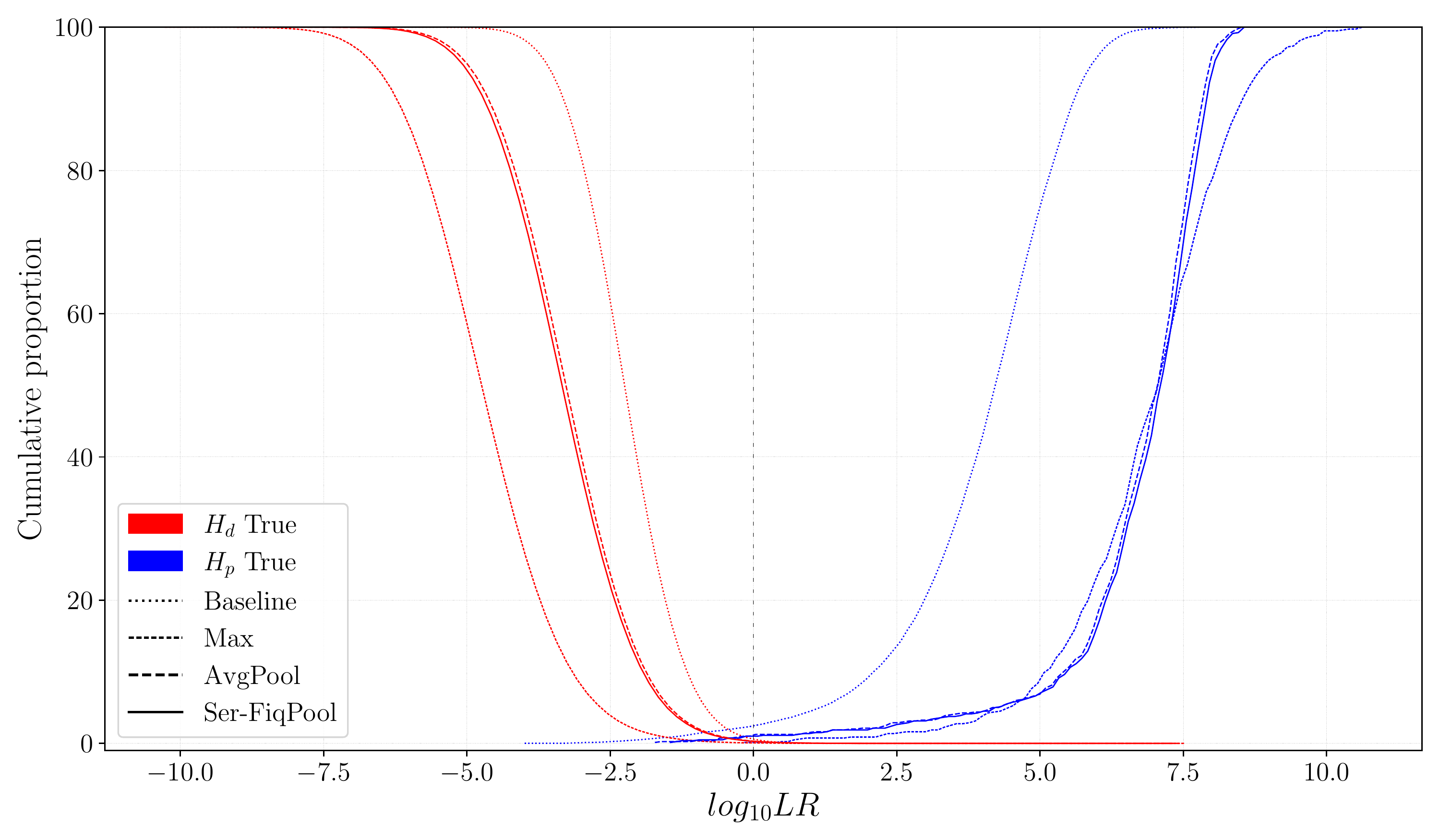}
         \caption{BFW cleaned}
     \end{subfigure}
        
        \caption{Tippett plots for the datasets of the social media scenario. Tippett plots for some strategies are omitted for clarity in the figure. Individual plots for every strategy are provided in the supplementary material file.}
        \label{fig:tippet_socialmedia}
\end{figure*}

Results for the social media scenario are shown in Table \ref{tab:cllr_social_media} and Figure \ref{fig:tippet_socialmedia}. We first note significant improvements in performance after cleaning both datasets. We also observe gains from the proposed aggregation strategies compared to the baselines. For some datasets, the \textit{MaxScore} strategy offered superior performance (lower \Cllr and greater separation of the Tippett plots) compared to the embedding aggregation strategies. We speculate this is due to the presence of images captured in the same session (e.g., consecutive frames of a video recording), which tends to produce very high scores when one of these images is selected as the reference but offers redundant information for the strategies that aggregate multiple embeddings. We do not investigate this possibility further since this work focuses on embedding aggregation, and both the \textit{AvgScore} and the \textit{MaxScore} are strategies based on score aggregation.

\begin{table}[h]
\centering
\caption{\Cllr for the social media scenario}\label{tab:cllr_social_media}
\resizebox{\columnwidth}{!}{
\begin{tabular}{|c| c| c| c| c|}
\hline
 & Adience 1 & Adience clean & BFW & BFW clean \\ 
\hline
Baseline    &0.174         &0.038         &0.217         &0.083 \\ 
AvgScore    &0.069         &0.008         &0.129         &0.036 \\ 
MaxScore    &\textbf{0.058}&0.010         &\textbf{0.088}&\textbf{0.003} \\ \hline 
AvgPool     &0.068         &0.007         &0.114         &0.027 \\ 
CSPool      &0.068         &\textbf{0.006}&0.114         &0.026 \\ 
Ser-FiqPool &0.068         &\textbf{0.006}&0.112         &0.025 \\ \hline
\end{tabular}}
\end{table}

In general, we observe that aggregating embeddings from multiple images of the same individual is an effective technique for improving recognition performance, especially of low-resolution images, which is especially interesting for the most challenging conditions in forensic casework. Even ``naïve''approaches such as \textit{AvgPool} can offer substantial performance improvements when the images are of similar quality. This strategy also has the advantage of not requiring the estimation of facial image quality.

\section{Conclusions}
\label{sec_conclusion}
We have presented approaches to improve face recognition performance under conditions usually found in forensic casework. We have demonstrated the benefits of these approaches to compute likelihood ratios derived from biometric scores. Although more sophisticated techniques for embedding aggregation exist, such as \citep{NAN, kim2022adaface}, we have demonstrated that even simple aggregation strategies may offer significant improvements for forensically realistic conditions.

As limitations of this work, we first note the absence of standard reference images in the social media datasets, with controlled pose, illumination, and facial expression. This is an important difference to forensic casework involving questioned-source images from social media. Also, the presence of multiple same-session images in these datasets makes it more difficult to generalize the results of score-based aggregation strategies (\textit{AvgScore} and \textit{MaxScore}) for casework. Regarding the surveillance scenario, the relatively small number of questioned-source images per identity and the similar quality of these images are also limiting factors compared to actual forensic data. We also acknowledge that the scores used in this work only consider the similarity of the facial images, disregarding their typicality. It has been shown  that this type of score is not ideal for computing LR under common-source hypotheses \citep{morrison_typicalitye_2018}.

We aim to address these limitations in future work by collecting new data and assessing the aggregation approaches on images from CCTV video that are more similar to casework conditions. We also aim to investigate more complex aggregation approaches based on neural networks, such as \citep{NAN, kim2022adaface}, and use scores that consider both similarity and typicality of the facial images.

 \bibliographystyle{elsarticle-num-names}
 \bibliography{cas-refs}





\end{document}